\documentclass{article}

\PassOptionsToPackage{numbers,compress}{natbib}
\usepackage[preprint]{neurips_2026}

\usepackage[utf8]{inputenc}
\usepackage[T1]{fontenc}
\usepackage{hyperref}
\usepackage{url}
\usepackage{booktabs}
\usepackage{amsfonts}
\usepackage{nicefrac}
\usepackage{microtype}
\usepackage{algorithm}
\let\AND\relax
\usepackage{algorithmic}
\usepackage{float}
\usepackage{graphicx}
\usepackage{subcaption}
\usepackage{multirow}
\usepackage[table]{xcolor}
\usepackage{amsmath}
\usepackage{bm}
\definecolor{cadcolor}{HTML}{2CA02C}
\definecolor{realcolor}{HTML}{D62728}

\title{Mitigating Performance Discrepancy in Cross-Domain 3D Class-Incremental Learning}

\author{%
  Jinge Ma \\
  Purdue University \\
  \texttt{ma859@purdue.edu}
  \And
  Gautham Vinod \\
  Purdue University \\
  \texttt{gvinod@purdue.edu}
  \And
  Bruce Coburn \\
  Purdue University \\
  \texttt{coburn6@purdue.edu}
  \AND
  Jui-Feng Chi \\
  Purdue University \\
  \texttt{chi96@purdue.edu}
  \And
  Siddeshwar Raghavan \\
  Purdue University \\
  \texttt{raghav12@purdue.edu}
  \And
  Fengqing Zhu \\
  Purdue University \\
  \texttt{zhu0@purdue.edu}
}

\begin{document}

\maketitle

\begin{abstract}
3D perception plays a crucial role in real-world applications such as autonomous driving, robotics, and AR/VR. In practical scenarios, 3D perception models need to continually adapt to newly emerging 3D object categories, making class-incremental learning (CIL) particularly important. However, unlike 2D images, 3D point clouds are inherently heterogeneous: objects from the same class may not only come from the clean CAD domain, but also from RGB-D camera scans of varying quality, video reconstructions, or even corrupted observations. We discover that such heterogeneity introduces a new challenge beyond catastrophic forgetting: the degree of performance degradation can vary substantially across domains, a phenomenon we term \textbf{performance discrepancy}. To investigate this problem, we establish the \textbf{Domain3D-CIL} training and evaluation protocol, which contains point cloud categories from heterogeneous domains. We further adapt a wide range of mainstream CIL methods to the 3D modality. The results demonstrate that this performance discrepancy consistently appears across these baselines. To mitigate this issue, we introduce \textbf{PolyMem}, an exemplar-free approach that implicitly models rich high-order statistics of the feature distribution to enhance cross-domain robustness. Experiments demonstrate that our method effectively alleviates the performance discrepancy while improving the model's performance across domains. Code will be made publicly available upon acceptance.
\end{abstract}

\section{Introduction}
3D perception holds broad application value in real-world scenarios~\cite{lang2019pointpillars,qi2019deep, chen2019cooper}. For example, autonomous driving systems require real-time recognition and classification of surrounding obstacles, while robots rely on 3D perception to understand and interact with complex physical environments. In practical deployments, 3D models must possess the capability to continually learn under constantly shifting data distributions. Class-Incremental Learning (CIL), a highly regarded branch in the field of continual learning, aims to enable models to incrementally learn new classes under the restricted condition where the historical data of old categories cannot be retained, while avoiding catastrophic forgetting of previously acquired knowledge~\cite{rebuffi2017icarl,li2017learning}. Catastrophic forgetting has been extensively studied in 2D CIL tasks, and it remains a core challenge for 3D CIL~\cite{chowdhury2022few}.

\begin{figure}[t]
    \centering
    \captionsetup[subfigure]{justification=centering}

    \begin{subfigure}[b]{0.328\linewidth}
        \centering
        \includegraphics[width=\linewidth]{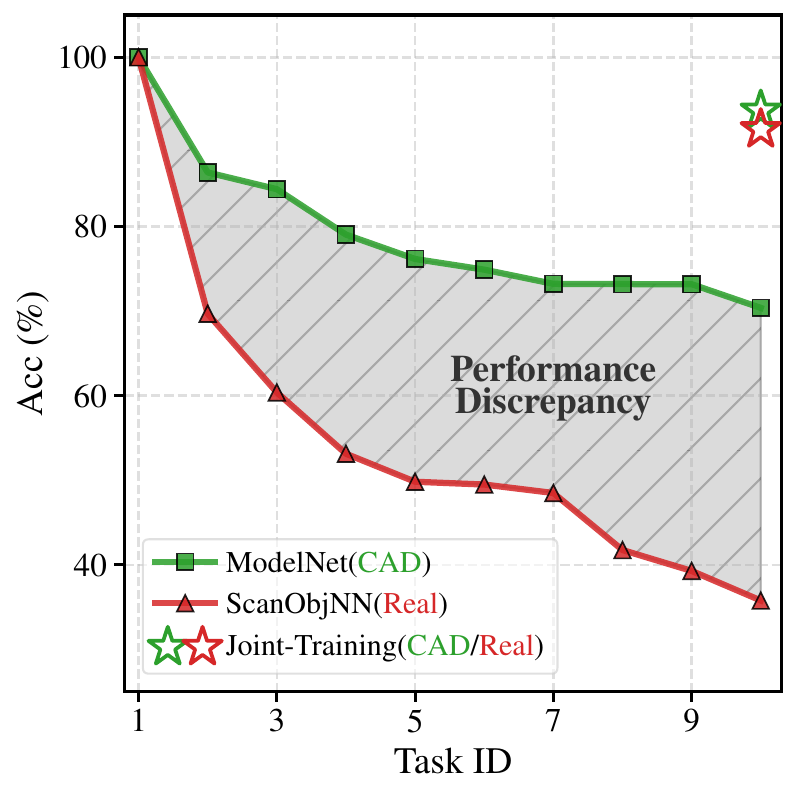}
        \caption{}
        \label{fig:fig1_a}
    \end{subfigure}%
    \begin{subfigure}[b]{0.33\linewidth}
        \centering
        \includegraphics[width=\linewidth]{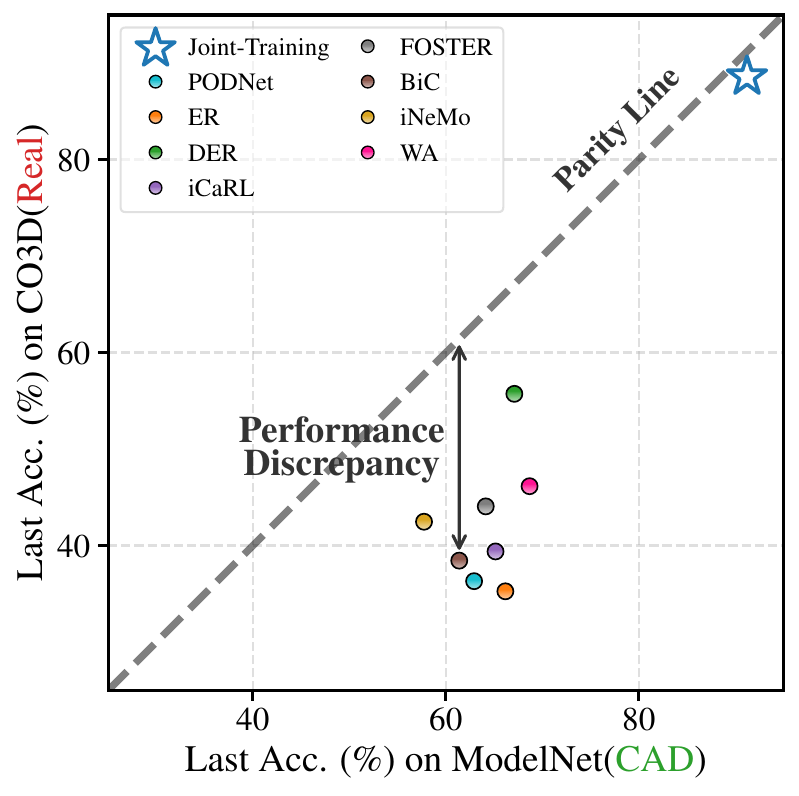}
        \caption{}
        \label{fig:fig1_b}
    \end{subfigure}
    \begin{subfigure}[b]{0.33\linewidth}
        \centering
        \includegraphics[width=\linewidth]{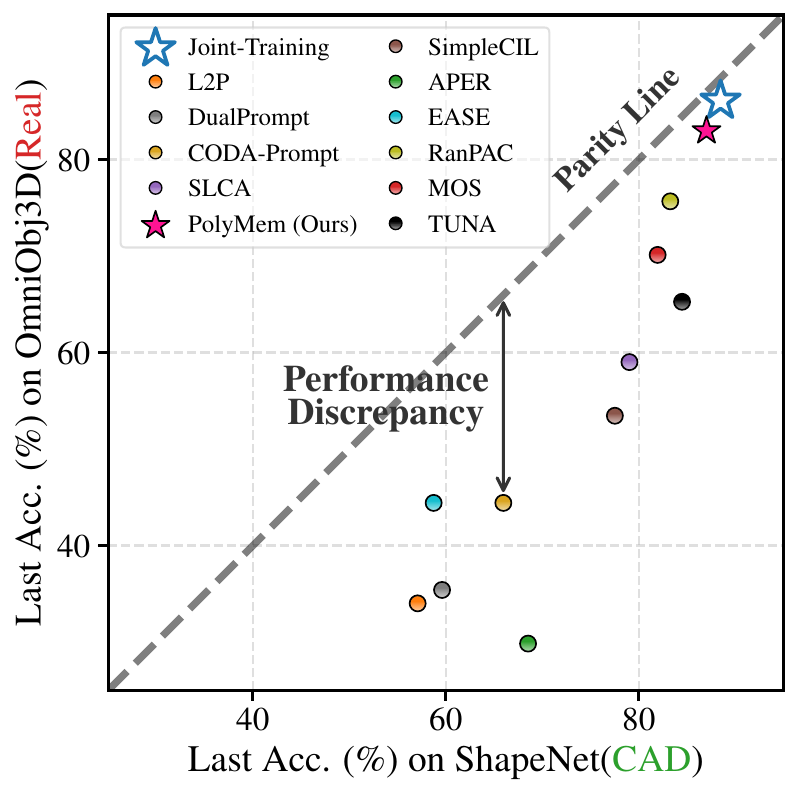}
        \caption{}
        \label{fig:fig1_c}
    \end{subfigure}

    \caption{\textbf{Visualization of performance discrepancy in 3D CIL.} \textbf{(a)} Illustrates the performance discrepancy of APER~\cite{zhou2025revisiting} between \textcolor{cadcolor}{CAD} and \textcolor{realcolor}{real}-scanned point clouds. As tasks progress, the performance gap between the CAD and the real-world domain gradually widens. In contrast, on Task 1 or under the joint-training setting, the performance across both domains is quite comparable. This demonstrates that the primary cause of the performance discrepancy is not static disparities in learning difficulty or in sample counts across domains. \textbf{(b)} and \textbf{(c)} demonstrate the prevalence of performance discrepancy across various baselines and different mixed datasets.}
    \label{fig:fig1}
    \vspace{-4mm}
\end{figure}

However, the inherent intra-class heterogeneity of 3D modalities poses unique challenges for CIL. Unlike 2D images primarily captured by cameras, the data sources for 3D objects are more diverse. Data may be sources from Computer-Aided Design (CAD) meshes, various RGB-D 3D scanners, video 3D reconstruction techniques, and so on~\cite{wu20153d,dai2017scannet,sun2020scalability}. Furthermore, due to the complexity of 3D environments, real-world point clouds are highly susceptible to containing corrupted or artifact-laden samples~\cite{ren2022benchmarking}. In the context of CIL, these factors cause point clouds of the same category to often be a mixture of samples from heterogeneous domains~\cite{qin2019pointdan}, thereby triggering a severe imbalance in memory retention across different domains. As shown in Fig.~\ref{fig:fig1_a}, as the number of tasks increases, the model not only exhibits overall catastrophic forgetting, but the degree of the model's performance degradation also presents a significant gap between different domains. We formally define this phenomenon, where the performance structurally diverges across heterogeneous domains, as \textbf{performance discrepancy}, which serves as the primary focus of this paper.

To systematically quantify and analyze the aforementioned challenges, we establish a comprehensive training and evaluation protocol, \textbf{Domain3D-CIL}. Under this protocol, the dataset consists of heterogeneous mixtures of clean CAD point clouds and real-world or corrupted point clouds. Samples from different data domains are mixed within shared semantic classes and share the same semantic labels. Detailed descriptions of Domain3D-CIL are provided in Sec.~\ref{sec:domain3d}. Regarding the baselines, we adopt the widely used Pre-Trained Model (PTM)-based CIL paradigm~\cite{zhou2024continual}, and systematically adapt representative methods under this paradigm to the 3D modality for fair comparison.

As shown in Fig.~\ref{fig:fig1_b} and Fig.~\ref{fig:fig1_c}, experiments under Domain3D-CIL confirm that performance discrepancy is a prevalent phenomenon among existing baselines. We find that some PTM-based baselines achieve strong performance by memorizing low-order statistics of old classes(e.g., means or covariances)~\cite{zhou2025revisiting,mcdonnell2023ranpac}. Furthermore, we find that the performance discrepancy of these baselines with statistical memory stems from the inherent limitation of using low-order statistics to represent class distributions in the feature space: low-order statistics fail to fully characterize all statistical properties, while many cross-domain differences are contained in higher-order statistics. To this end, we introduce \textbf{PolyMem}. PolyMem leverages a structured polynomial approximation of the RBF kernel to capture higher-order feature interactions, thereby achieving robust learning for cross-domain 3D data. Evaluations show that PolyMem effectively mitigates performance discrepancy across heterogeneous 3D domains, while improving performance within each domain.

In summary, the main contributions of this paper are as follows:

\begin{itemize}
    \item We formalize the phenomenon of \textbf{performance discrepancy} in 3D CIL, and establish the \textbf{Domain3D-CIL} protocol with heterogeneous 3D data to simulate real-world scenarios.
    \item We transfer widely used CIL baselines to the 3D modality, and analyze the mechanisms causing performance discrepancy of some baselines under cross-domain settings.
    \item We propose the \textbf{PolyMem} framework. Our method effectively mitigates cross-domain performance discrepancy as well as within-domain catastrophic forgetting.
\end{itemize}

\section{Related Works}

\paragraph{3D Representation Learning and Data Heterogeneity.}
The evolution of 3D point cloud representation learning has played a crucial role in modern 3D perception. Early representative works, such as PointNet and PointNet++, established a point-set processing framework based on shared multi-layer perceptrons~\cite{qi2017pointnet,qi2017pointnet++}. In recent years, Transformer-based architectures have further improved 3D representation learning through more effective global context modeling~\cite{zhao2021point,wu2022point,wu2024point}. Building on these architectural advances, research has gradually shifted toward large-scale pre-trained models. Works such as Point-BERT have established self-supervised pre-training paradigms for 3D point clouds through masked point modeling~\cite{yu2022point,pang2023masked,zhang2022point}. On this basis, methods such as Uni3D further leverage multimodal alignment across text, images, and point clouds to learn semantically consistent 3D representations~\cite{xue2023ulip,liu2023openshape,xue2024ulip,zhouuni3d}.

As recent studies explore unified representation learning over heterogeneous 3D data sources~\cite{wu2025sonata,zhang2025concerto}, 3D representation learning is extending to real-world scenarios, where data heterogeneity becomes increasingly prominent. Differences in data construction and acquisition processes (e.g., CAD modeling~\cite{wu20153d,chang2015shapenet}, RGB-D scanning~\cite{uy2019revisiting,dai2017scannet,wu2023omniobject3d}, LiDAR acquisition~\cite{geiger2013vision,sun2020scalability}, and multi-view reconstruction~\cite{reizenstein2021common,yu2023mvimgnet}) introduce significantly different domain properties. As a result, samples from the same semantic class across datasets exhibit highly complex intra-class distributions in the feature space, with substantial variations in representativeness, diversity, noise patterns, and feature distributions, etc. This trend highlights the growing need for more robust 3D representations in complex real-world environments.

\paragraph{Class-incremental Learning.}
Class-incremental learning aims to continuously learn new classes while retaining knowledge of previously learned ones~\cite{zhou2024continual}. In the CIL setting, models are trained on a sequence of tasks, each introducing new classes, without access to full historical data. Early approaches address this challenge through exemplar replay~\cite{rebuffi2017icarl,wu2019large,hou2019learning,ma2026temporal}, parameter regularization~\cite{li2017learning,kirkpatrick2017overcoming,miao2021continual}, or architectural expansion~\cite{yan2021dynamically,wang2022foster,zhou2022model} to mitigate inter-task interference. Although these strategies differ in implementation, strong baselines often still rely on storing replay exemplars from previous classes. Building on this line of research, Pre-Trained Model (PTM)-based CIL has recently emerged as an important paradigm~\cite{zhou2024continual}. These methods leverage a pre-trained backbone with strong generalization and typically follow a replay-free protocol, where no raw data from old classes are stored. The backbone is usually frozen, and incremental adaptation is achieved through parameter-efficient modules. Representative approaches include prompt-based strategies~\cite{wang2022learning,wang2022dualprompt,smith2023coda} as well as methods that enhance adaptation via feature calibration~\cite{zhang2023slca,goswami2023fecam}, lightweight mappings~\cite{gao2023unified,sun2025mos}, or subspace expansion~\cite{zhou2024expandable,wang2025integrating,zhou2025dual}. In the replay-free setting, some methods form statistical memory by recording feature-space statistics. Specifically, SimpleCIL~\cite{zhou2025revisiting} only retains the 1st-order statistics of feature distributions, while RanPAC~\cite{mcdonnell2023ranpac} further combines 1st-order with 2nd-order feature statistics.

In the 3D domain, existing incremental learning studies have mainly focused on few-shot settings~\cite{chowdhury2022few,cheraghian2024canonical,xiang2025seeing}. Prior works have explored 2D domain-incremental~\cite{wang2022s,wang2024non} or class-and-domain incremental learning~\cite{xie2022general,park2024versatile}, but they mostly focus on domain shifts across tasks, while the cross-domain performance discrepancy induced by heterogeneous 3D data mixed within the same semantic class remains underexplored. In this work, we focus on analyzing this cross-domain performance discrepancy under the PTM-based CIL paradigm in heterogeneous 3D environments.

\section{Preliminaries}
\subsection{Class-incremental Learning}
\label{sec:prelim}
We consider a sequence of class-incremental tasks $\{\mathcal{T}_t\}_{t=1}^T$. The training set of task $t$ is denoted by $\mathcal{D}_t=\{(x_{t,i},\,y_{t,i})\}_{i=1}^{n_t}$, where $x_{t,i}$ and $y_{t,i}$ denote the $i$-th sample and its label in task $t$, respectively, $n_t$ is the number of training samples in task $t$, and $\mathcal{Y}_t$ denotes the label set of task $t$. In the CIL setting, the label sets of different tasks are mutually disjoint, i.e., $\mathcal{Y}_t \cap \mathcal{Y}_{t'}=\emptyset$ for any $t \neq t'$. Let $f_t$ denote the model obtained after learning task $t$, which is evaluated on all classes observed so far, i.e., on $\bigcup_{\tau=1}^{t} \mathcal{Y}_\tau$. Since the full datasets from previous tasks are no longer accessible in later stages, learning new tasks typically leads to catastrophic forgetting on old classes. Following the PTM-based CIL paradigm, the initial model $f_0$ is built upon a pre-trained transformer-based backbone and follows a replay-free setting, where the model is trained at each stage using only the current task data $\mathcal{D}_t$ without storing raw samples from previous classes. In this work, we instantiate this paradigm in the 3D domain using \textbf{Uni3D-Base}~\cite{zhouuni3d} pre-trained on Objaverse~\cite{deitke2023objaverse} as the backbone. Uni3D is adopted because it shares a transformer-based architectural design with Vision Transformer~\cite{dosovitskiy2021an}, which facilitates the adaptation of existing PTM-based CIL methods to 3D perception tasks.

\subsection{Domain3D-CIL}
\label{sec:domain3d}
\begin{figure}[t]
    \centering
    \begin{subfigure}[b]{0.65\linewidth}
        \centering
        \includegraphics[width=\linewidth]{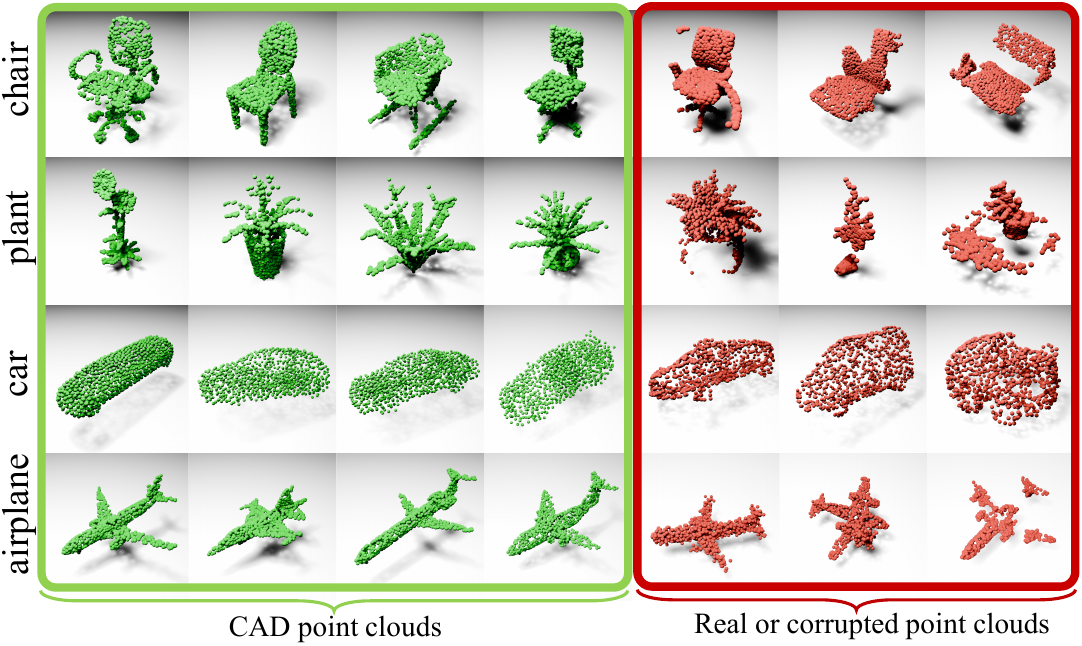}
        \caption{}
        \label{fig:fig2_a}
    \end{subfigure}%
    \hspace{0.02\linewidth}%
    \begin{subfigure}[b]{0.31\linewidth}
        \centering
        \raisebox{0cm}{\includegraphics[width=\linewidth]{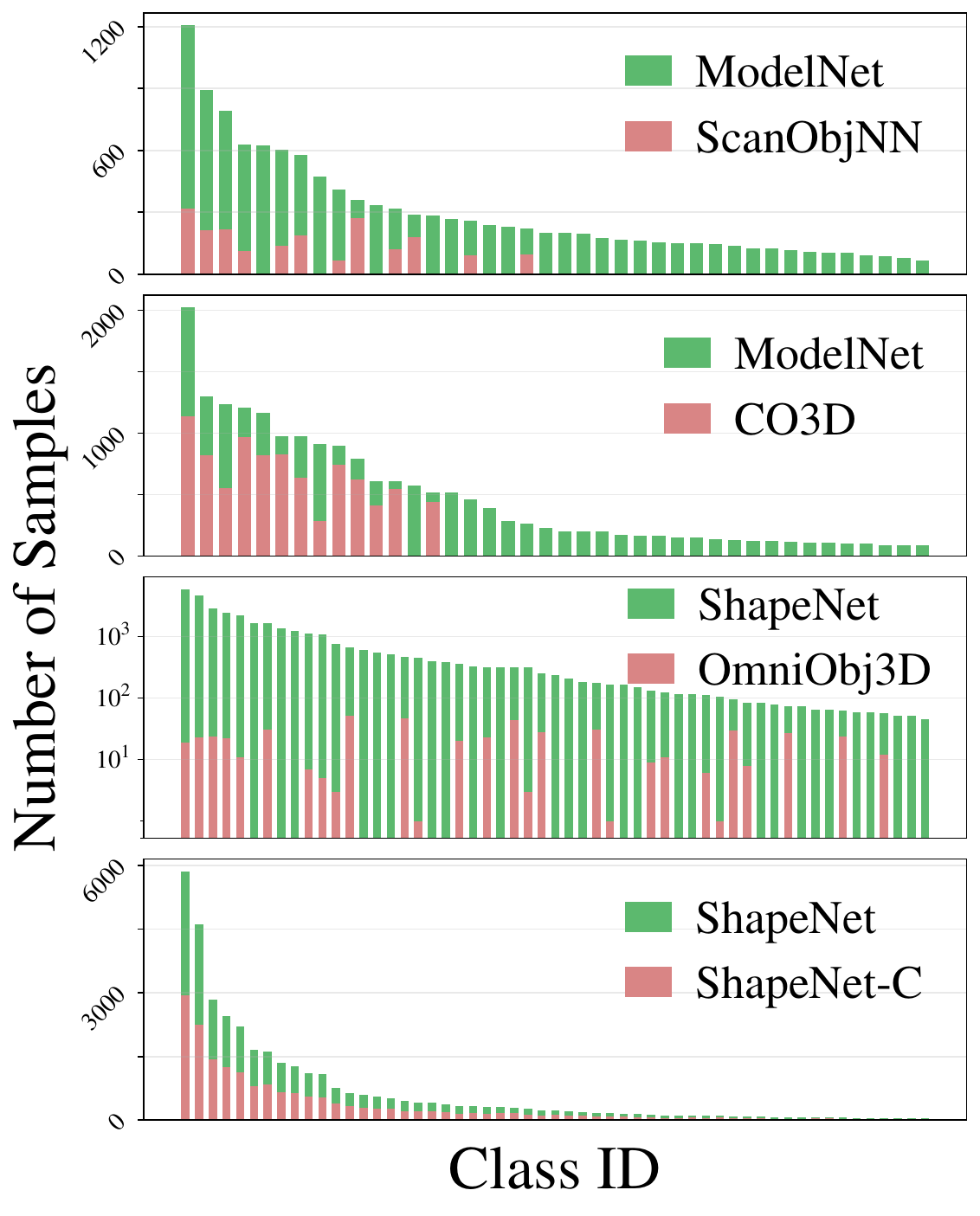}}
        \caption{}
        \label{fig:fig2_b}
    \end{subfigure}
    \caption{\textbf{Visualization of Domain3D-CIL.}
    \textbf{(a)} shows representative categories containing cross-domain samples from the mixed datasets.
    \textbf{(b)} shows the corresponding class distributions.}
    \label{fig:fig2}
\end{figure}
\vspace{-2mm}

Building upon the general CIL formulation above, we further establish a heterogeneous 3D training and evaluation protocol, termed \textbf{Domain3D-CIL}. For each task $t$, the training set is constructed by mixing samples from a canonical CAD domain and one additional domain, i.e., $\mathcal{D}_t = \mathcal{D}_t^{\mathrm{CAD}} \cup \mathcal{D}_t^{\mathrm{O}}$, where $\mathcal{D}_t^{\mathrm{CAD}}=\{(x_{t,i}^{\mathrm{CAD}},y_{t,i}^{\mathrm{CAD}})\}_{i=1}^{n_t^{\mathrm{CAD}}}$ and $\mathcal{D}_t^{\mathrm{O}}=\{(x_{t,j}^{\mathrm{O}},y_{t,j}^{\mathrm{O}})\}_{j=1}^{n_t^{\mathrm{O}}}$.\vadjust{\vspace{0.5ex}} Here, the additional domain $\mathrm{O}$ is chosen from one of the following settings: real-world RGB-D point clouds, video-reconstructed point clouds, or corrupted CAD point clouds. In each experimental setting, only one such non-CAD domain is mixed with the CAD domain. For each task, the CAD subset is constructed over the full task label space $\mathcal{Y}_t$, while the non-CAD subset only contains samples that can be semantically matched to CAD categories. Let $\mathcal{Y}_t^{\mathrm{O}}$ denote the label space of the non-CAD subset, then $\mathcal{Y}_t^{\mathrm{O}}\subseteq \mathcal{Y}_t$. Samples with the same semantic meaning across different domains share the same class label and are not treated as new domain-specific categories. Non-CAD categories that cannot be matched to the current CAD label space are not included, while CAD categories without corresponding non-CAD samples are still retained in $\mathcal{D}_t^{\mathrm{CAD}}$.

During training, the model only observes the mixed training set $\mathcal{D}_t$ together with semantic labels, while domain labels are \textbf{not provided}. During evaluation, we split the test data into domain-specific subsets, such as $\mathcal{D}_{t,\mathrm{test}}^{\mathrm{CAD}}$ and $\mathcal{D}_{t,\mathrm{test}}^{\mathrm{O}}$, and report performance separately on each domain over all observed classes $\mathcal{C}_{t-1}=\bigcup_{\tau=1}^{t}\mathcal{Y}_\tau$. This design is based on two considerations. First, domain boundaries in real-world 3D data are often continuous and ambiguous; differences in acquisition devices, sampling conditions, and preprocessing pipelines can all introduce distribution shifts, making discrete domain labels difficult to define reliably. Second, 3D data in real applications may be collected from the web and usually lacks explicit domain labels. In this way, Domain3D-CIL provides a unified protocol for systematically evaluating both average performance and domain-wise performance discrepancy under heterogeneous 3D CIL data streams.

Fig.~\ref{fig:fig2} shows the mixed point cloud datasets constructed for experiments under Domain3D-CIL. Each row in Fig.~\ref{fig:fig2_a} and Fig.~\ref{fig:fig2_b} corresponds to the same mixed dataset, whose composition is indicated in the legend of Fig.~\ref{fig:fig2_b}. From top to bottom, the four mixed datasets are ModelNet~\cite{wu20153d} + ScanObjectNN~\cite{uy2019revisiting}, ModelNet + CO3D~\cite{reizenstein2021common}, ShapeNet~\cite{chang2015shapenet} + OmniObject3D~\cite{wu2023omniobject3d}, and ShapeNet + ShapeNet-C~\cite{ren2022benchmarking}. As shown in Fig.~\ref{fig:fig2_a}, CAD point clouds such as ModelNet and ShapeNet are generally clean and complete, and high-quality RGB-D point clouds from OmniObject3D also exhibit relatively clear geometry. In contrast, low-quality RGB-D scans from ScanObjectNN, video-reconstructed point clouds from CO3D, and corrupted point clouds from ShapeNet-C contain more artifacts, including missing regions, redundant points, and perturbations. Fig.~\ref{fig:fig2_b} further shows the class distributions of these mixed datasets, all of which exhibit clear long-tail characteristics. In particular, the 3rd row uses a logarithmic y axis, highlighting that the sample scale of OmniObject3D is much smaller than that of ShapeNet, since high-quality RGB-D point clouds are harder to collect and thus scarcer than CAD models.

\section{Methodology}
\subsection{Statistical Memory in PTM-based CIL}
\label{sec:ptm-based}

As shown in Fig.~\ref{fig:fig1_a} and Fig.~\ref{fig:fig1_c}, we observe a cross-domain performance discrepancy of the PTM-based CIL baselines and the Domain3D-CIL protocol. In this section, we further investigate the underlying causes of this phenomenon. PTM-based CIL baselines benefit from powerful pre-trained backbones. Since their feature mappings can remain frozen and strictly stationary, some baselines can effectively represent and persistently memorize old classes by directly recording feature-space statistics, even without relying on raw exemplar replay. We refer to this as \textbf{statistical memory}.

The most direct form of statistical memory is given by \textbf{SimpleCIL}~\cite{zhou2025revisiting}.
Given a frozen pre-trained backbone, each sample $x_i$ is mapped to a stable feature vector
$F_i=f(x_i)\in\mathbb{R}^D$.
For each class $c$, SimpleCIL summarizes the class distribution by its prototype, namely the class-wise mean in the feature space:
\vspace{-3mm}
\begin{equation}
\bm{\mu}_c=\frac{1}{N_c}\sum_{i:y_i=c} F_i,
\quad \bm{\mu}_c\in\mathbb{R}^{D},
\end{equation}
where $N_c$ denotes the number of samples in class $c$, and $\bm{\mu}_c$ denotes the prototype vector of class $c$.
During inference, classification is performed by a prototype-based rule, equivalently by using these prototypes as classifier weights or by applying a nearest-mean decision rule.
Therefore, SimpleCIL preserves only the \textbf{1st-order statistic} (mean) of each class distribution in the feature space.
Although this representation is mathematically concise, it only retains the 1st-order statistic and cannot explicitly characterize \textbf{2nd-order} properties. It effectively treats all class distributions as having identical covariance and therefore fails to reflect intra-class diversity.

\begin{figure}[h]
    \centering
    \begin{subfigure}[b]{0.242\linewidth}
        \centering
        \raisebox{0.04cm}{\includegraphics[width=\linewidth]{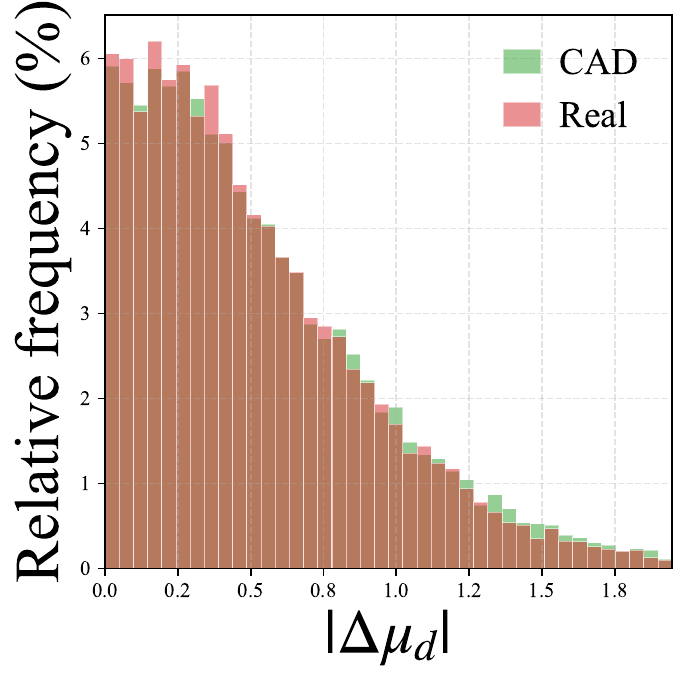}}
        \caption{1st: Mean}
        \label{fig:fig3_a}
    \end{subfigure}%
    \hspace{0.01\linewidth}%
    \begin{subfigure}[b]{0.24\linewidth}
        \centering
        \raisebox{0cm}{\includegraphics[width=\linewidth]{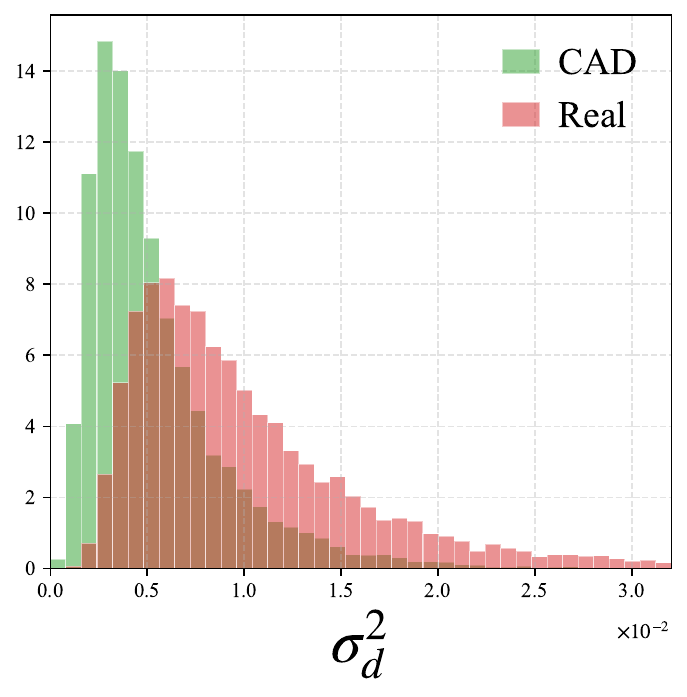}}
        \caption{2nd:Variance}
        \label{fig:fig3_b}
    \end{subfigure}
    \begin{subfigure}[b]{0.243\linewidth}
        \centering
        \raisebox{0.065cm}{\includegraphics[width=\linewidth]{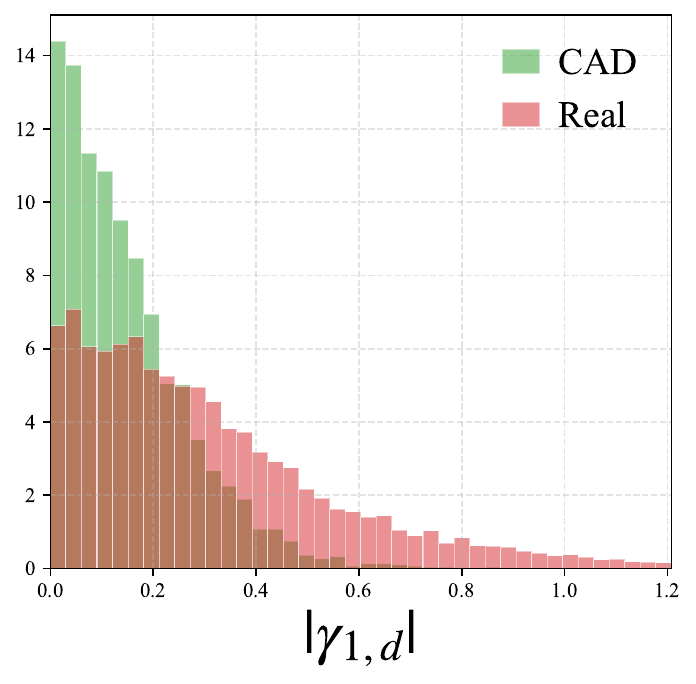}}
        \caption{3rd:Skewness}
        \label{fig:fig3_c}
    \end{subfigure}
    \begin{subfigure}[b]{0.24\linewidth}
        \centering
        \raisebox{0.07cm}{\includegraphics[width=\linewidth]{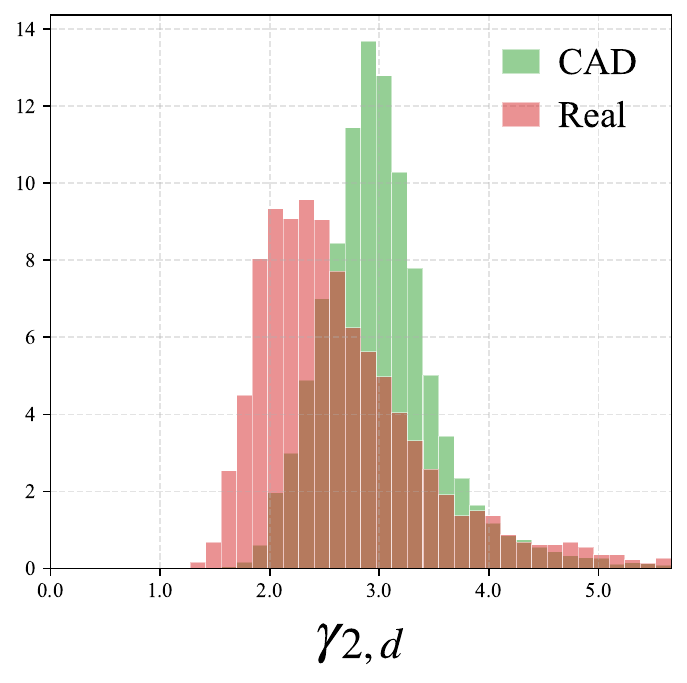}}
        \caption{4th:Kurtosis}
        \label{fig:fig3_d}
    \end{subfigure}
    \caption{Histograms of 1st to 4th-order statistics on ShapeNet(\textcolor{cadcolor}{CAD})+OmniObject3D(\textcolor{realcolor}{Real}).}
    \label{fig:fig3}
\end{figure}

We verify the failure mode of SimpleCIL under Domain3D-CIL by visualizing the feature space of the backbone.
Fig.~\ref{fig:fig3} shows the distributional differences of 1st- to 4th-order feature statistics between CAD and real domains on the mixed dataset of ShapeNet+OmniObject3D. We extract features from the two domains and compute the corresponding per-dimension statistics along the feature dimensions ($d\in\{1,\dots,1024\}$). For the 1st-order statistic, we compute the class mean within each domain and measure its per-dimension deviation from the overall class prototype $\bm{\mu}_c$, which is denoted as $|\Delta\mu_d|$ in Fig.~\ref{fig:fig3_a}.
For the remaining statistics, we compute the per-dimension variance $\sigma_d^2$, absolute skewness $|\gamma_{1,d}|$, and kurtosis $\gamma_{2,d}$. The relative-frequency histograms of these statistics are averaged over 10 semantically matched classes. According to Fig.~\ref{fig:fig3_a}, the CAD and real domains exhibit similar 1st-order statistical distributions, indicating that the per-dimension deviations of the domain-specific class means from the prototype are comparable, with no obvious domain gap.
However, Fig.~\ref{fig:fig3_b} shows a clear difference in 2nd-order statistics: more feature dimensions in the CAD domain are concentrated in the low-variance range, whereas the real domain contains more high-variance dimensions, suggesting stronger intra-class diversity in real scanned objects. Such diversity cannot be captured by 1st-order statistics alone, making real point clouds more vulnerable to interference from new-task prototypes and thereby contributing to performance discrepancy.

A stronger form of statistical memory is introduced by \textbf{RanPAC}~\cite{mcdonnell2023ranpac},
which extends the 1st-order memory by preserving both \textbf{1st- and 2nd-order} statistics in the lifted feature space through the accumulated statistics of \textbf{ridge regression}~\cite{hoerl1970ridge}:
\begin{equation}
Q=\sum_{i=1}^{N}F_i' e_{y_i}^\top,\qquad
G=\sum_{i=1}^{N}F_i' {F_i'}^\top.
\label{Eq:ridge_regression}
\end{equation}
Here, $e_{y_i}\in\mathbb{R}^{C}$ denotes the one-hot label vector of sample $x_i$.
The lifted feature of sample $i$ is defined as
$F_i' := \phi(F_i)=\phi(f(x_i)) \in \mathbb{R}^M$,
where $\phi:\mathbb{R}^D\rightarrow\mathbb{R}^M$ denotes a random nonlinear mapping.
Specifically, $Q \in \mathbb{R}^{M \times C}$ aggregates 1st-order statistics in the lifted feature space; $G \in \mathbb{R}^{M \times M}$ stores a global uncentered \textbf{2nd-order} statistic, which characterizes the 2nd-order distributional structure in the lifted feature space and is closely related to the covariance structure.
Based on these two statistics, RanPAC learns a linear predictor $s(x_i)=W^\top F_i'\in\mathbb{R}^C$, where the classifier is obtained by ridge regression: $W=(G+\lambda I)^{-1}Q.$ Here, $\lambda$ is the ridge coefficient selected from a fixed grid using a validation split on the current task, and $I$ is the identity matrix. The additional 2nd-order memory can better reflect intra-class diversity in Fig.~\ref{fig:fig3_b}, thereby alleviating the performance discrepancy caused by the absence of 2nd-order statistics in SimpleCIL.
This also explains why RanPAC usually exhibits a smaller performance discrepancy and stronger overall performance than other baselines in Sec.~\ref{sec:main_exp}.

\subsection{Proposed Method: PolyMem}
\label{sec:PolyMem}

As shown in Fig.~\ref{fig:fig1_c}, the improvement of RanPAC over SimpleCIL suggests that incorporating 2nd-order statistics can alleviate cross-domain performance discrepancy. Moreover, RanPAC constructs an expanded feature space through random nonlinear projection, thereby implicitly introducing nonlinear feature interactions. However, such a randomly lifted space remains heuristic: although it may produce nonlinear interactions, it does not provide a structured, order-wise interpretation of how these interactions correspond to different orders of statistical structures. That is, the lifted representation in RanPAC is not explicitly tied to a theoretically grounded order-wise feature space.

Based on this observation, a natural but naive extrapolation is to ask: if we could further memorize higher-order statistics of old classes, could we more faithfully characterize the 3D domain gap in complex feature distributions? For example, skewness (\textbf{3rd-order} statistic) reflects the asymmetry of a distribution~\cite{pearson1895skewness}. In Fig.~\ref{fig:fig3_c}, the real domain has larger absolute skewness in more feature dimensions. A possible explanation is that artifacts in real scanned point clouds are not random perturbations, but repeatedly appear along similar feature directions, thereby inducing structural asymmetry in the feature space. In Fig.~\ref{fig:fig3_d}, kurtosis (\textbf{4th-order} statistic) reflects the heavy-tailed behavior of a distribution~\cite{pearson1905rejoinder}, and ShapeNet exhibits a more evident heavy-tailed effect than OmniObject3D.

Although statistical memory eliminates the overhead of storing all raw features, explicitly storing higher-order statistics of feature distributions inevitably requires additional memory. For example, for a feature distribution in $\mathbb{R}^D$, explicitly recording its $k$-th order statistic requires maintaining a tensor with $\mathcal{O}(D^k)$ elements. As the recorded order $k$ increases, the required storage capacity grows \textbf{exponentially}. Polynomial interactions are natural feature-level building blocks for such high-order statistics. To avoid explicit storage of high-order statistics while moving beyond the heuristic random projection in RanPAC, we propose \textbf{PolyMem}. PolyMem replaces the random projection with a structured polynomial approximation of the RBF kernel~\cite{scholkopf2002learning}, following the kernel approximation framework~\cite{cui2017kernel}. This framework was originally developed for fine-grained classification with end-to-end CNN training. We adapt this construction to PTM-based CIL: the lifted features are accumulated over a frozen pre-trained backbone for closed-form ridge regression, with kernel coefficients fixed at the RBF Taylor values rather than learned via backprop. In this way, PolyMem injects order-wise higher-order interactions into statistical memory, without storing high-order statistical tensors.

Specifically, let $F_i, F_j \in \mathbb{R}^D$ denote two L2-normalized feature vectors in the backbone feature space. Rather than relying on an random lifted space, we seek a more structured feature space induced by a principled similarity prior. To this end, we consider the RBF kernel:
\begin{equation}
k_{\mathrm{RBF}}(F_i, F_j)=\exp\!\left(-\gamma \|F_i-F_j\|_2^2\right)=
\exp(-\gamma\|F_i\|_2^2)\exp(-\gamma\|F_j\|_2^2)\exp(2\gamma F_i^\top F_j).
\end{equation}
We set $\gamma$ by the median heuristic on normalized features. The last factor admits the Taylor expansion:
\begin{equation}
\begin{aligned}
\exp(2\gamma F_i^\top F_j)=
1
+
\underbrace{2\gamma(F_i^\top F_j)}_{\text{1st-order}}
+
\underbrace{\frac{(2\gamma)^2}{2!}(F_i^\top F_j)^2}_{\text{2nd-order}}
+
\underbrace{\frac{(2\gamma)^3}{3!}(F_i^\top F_j)^3}_{\text{3rd-order}}
+\cdots .
\end{aligned}
\end{equation}

This expansion shows that the RBF kernel can be decomposed into polynomial feature interactions of different orders. The linear term captures 1st-order interactions between the two feature vectors, the quadratic term captures 2nd-order interactions, and higher-degree terms further encode increasingly complex feature interactions. Therefore, the RBF expansion provides a natural order-wise form for constructing structured feature representations.

To obtain a controllable order-wise representation in the \textbf{mapped} feature space, PolyMem retains the Taylor expansion up to order $L$.
For original features, this corresponds to the following finite-order polynomial kernel:
\vspace{-3mm}
\begin{equation}
k_{\mathrm{Poly}}^{(L)}(F_i,F_j)
=
e^{-2\gamma}
\sum_{n=0}^{L}
\frac{(2\gamma)^n}{n!}
(F_i^\top F_j)^n .
\end{equation}
Directly constructing the corresponding polynomial features would still be expensive. Therefore, PolyMem adopts TensorSketch~\cite{pham2013fast} to approximate each polynomial term in a compact form. Let $\mathrm{TS}_n(F_i)\in\mathbb{R}^{M_n}$ denote the sketch feature for the $n$-th order interaction, where $1+\sum_{n=1}^{L}M_n=M$. The remaining dimensions are evenly allocated across  orders. Then the lifted feature is defined as
\begin{equation}
\phi_{\mathrm{Poly}}(F_i)
=
\left[
e^{-\gamma},
\;
e^{-\gamma}\sqrt{\frac{2\gamma}{1!}}\mathrm{TS}_1(F_i),
\;
\dots,
\;
e^{-\gamma}\sqrt{\frac{(2\gamma)^L}{L!}}\mathrm{TS}_L(F_i)
\right]
\in\mathbb{R}^{M}.
\end{equation}
This mapping satisfies$\phi_{\mathrm{Poly}}(F_i)^\top
\phi_{\mathrm{Poly}}(F_j)
\approx
k_{\mathrm{Poly}}^{(L)}(F_i,F_j)$, thereby approximating the first $L$ orders of feature interactions within a fixed dimension $M$. Following the notation in Sec.~\ref{sec:ptm-based}, we define the resulting lifted feature of sample $i$ as
$F_i' := \phi_{\mathrm{Poly}}(F_i) \in \mathbb{R}^M$.
PolyMem does not modify the underlying closed-form learning procedure. Instead, it directly substitutes this order-wise polynomial lifted feature into the same accumulated-statistics form underlying ridge regression. That is, by using $F_i'$ in Eq.~\ref{Eq:ridge_regression}, PolyMem accumulates $Q$ and $G$ in the mapped space without storing old samples, and obtains the classifier weights $W$ through the same closed-form ridge regression solution. \textbf{We provide the theoretical background in the Appendix.}

This design brings two key advantages. First, it replaces the heuristic random lifting in RanPAC with a structured order-wise polynomial mapping, making different orders of feature interactions explicit and analyzable. Second, it enables implicit high-order statistical modeling while retaining the same memory-efficient accumulated statistics. Although PolyMem still only accumulates $Q$ and $G$, these statistics are computed over polynomially lifted features and therefore reflect richer feature interactions of the original feature distribution up to order $L$. Consequently, PolyMem improves the distribution modeling capacity of statistical memory while preserving the replay-free and memory-efficient advantages of ridge regression.

\section{Experiments}

\subsection{Experimental Setup}
\paragraph{Baselines.}We adopt the following PTM-based CIL methods as baselines:
L2P~\cite{wang2022learning}, DualPrompt~\cite{wang2022dualprompt}, CODA-Prompt~\cite{smith2023coda}, SLCA~\cite{zhang2023slca},SimpleCIL~\cite{zhou2025revisiting}, APER~\cite{zhou2025revisiting}, EASE~\cite{zhou2024expandable}, RanPAC~\cite{mcdonnell2023ranpac}, MOS~\cite{sun2025mos}, and TUNA~\cite{wang2025integrating}.
For a fair comparison, as described in Sec.~\ref{sec:ptm-based}, these baselines use the same backbone.
\textbf{Datasets.}We use 6 widely used public 3D datasets to construct cross-domain mixed 3D data. Specifically, ModelNet~\cite{wu20153d} and ShapeNet~\cite{chang2015shapenet} belong to CAD domains, mainly consisting of clean 3D models collected from online sources.
ShapeNet-C~\cite{ren2022benchmarking} is a corrupted point cloud domain built upon ShapeNet, designed to simulate common 3D corruptions.
ScanObjectNN~\cite{uy2019revisiting} contains real indoor object point clouds scanned by low-cost RGB-D cameras.
CO3D~\cite{reizenstein2021common} is collected from crowdsourced smartphone videos, with real object point clouds reconstructed via Structure-from-Motion.
OmniObject3D~\cite{wu2023omniobject3d} contains high-quality real-scanned 3D objects across multiple categories.
The training and evaluation protocols follow Domain3D-CIL, as detailed in Sec.~\ref{sec:domain3d}.
\textbf{Evaluation Metrics.}
We report performance on two domains, denoted as \textbf{\textcolor{cadcolor}{$D_1$}} and \textbf{\textcolor{realcolor}{$D_2$}}, where $D_1$ corresponds to the CAD domain and $D_2$ corresponds to the real or corrupted domain. Following standard CIL evaluation, we adopt two base metrics: \textbf{Last Accuracy (LA)}, the accuracy over all classes after the final incremental task, i.e., $\mathrm{LA}=a_T$; and \textbf{Cumulative Accuracy (CA)}, the average accuracy over all incremental tasks, i.e., $\mathrm{CA}=\frac{1}{T}\sum_{t=1}^{T}a_t$. For each metric, we report the performance on $D_1$, on $D_2$, and their difference $D_1-D_2$. We formally define \textbf{performance discrepancy} as this non-absolute cross-domain metric gap, i.e., $Gap =\mathrm{Metric}_{D_1}-\mathrm{Metric}_{D_2}$, where $\mathrm{Metric}\in\{\mathrm{LA},\mathrm{CA}\}$. A smaller metric gap indicates better cross-domain balance. All reported results are averaged over three runs.

\subsection{Main Results}
\label{sec:main_exp}

\begin{table*}[!ht]
\centering
\caption{Quantitative evaluation of PTM-based CIL baselines and PolyMem on Domain3D-CIL.}
\label{tab:ptm_results_full}

\resizebox{1\linewidth}{!}{
\begin{tabular}{l ccc ccc ccc ccc}
\toprule
\multirow{3}{*}{Method}
  & \multicolumn{6}{c}{ModelNet (\textcolor{cadcolor}{$D_1$}) + ScanObjectNN (\textcolor{realcolor}{$D_2$})}
  & \multicolumn{6}{c}{ModelNet (\textcolor{cadcolor}{$D_1$}) + CO3D (\textcolor{realcolor}{$D_2$})} \\
\cmidrule(lr){2-7} \cmidrule(lr){8-13}
  & \multicolumn{3}{c}{Last Acc.}
  & \multicolumn{3}{c}{Cum. Acc.}
  & \multicolumn{3}{c}{Last Acc.}
  & \multicolumn{3}{c}{Cum. Acc.} \\
\cmidrule(lr){2-4} \cmidrule(lr){5-7} \cmidrule(lr){8-10} \cmidrule(lr){11-13}
  & \textcolor{cadcolor}{$D_1$}$\!\uparrow$ & \textcolor{realcolor}{$D_2$}$\!\uparrow$ & $\mathrm{Gap}\!\downarrow$
  & \textcolor{cadcolor}{$D_1$}$\!\uparrow$ & \textcolor{realcolor}{$D_2$}$\!\uparrow$ & $\mathrm{Gap}\!\downarrow$
  & \textcolor{cadcolor}{$D_1$}$\!\uparrow$ & \textcolor{realcolor}{$D_2$}$\!\uparrow$ & $\mathrm{Gap}\!\downarrow$
  & \textcolor{cadcolor}{$D_1$}$\!\uparrow$ & \textcolor{realcolor}{$D_2$}$\!\uparrow$ & $\mathrm{Gap}\!\downarrow$ \\
\midrule
Joint-Training
            & $\overline{93.68}$ & $\overline{91.48}$ & $\overline{2.20}$ & - & - & -
            & $\overline{92.88}$ & $\overline{89.21}$ & $\overline{3.67}$ & - & - & - \\
L2P~\cite{wang2022learning}
            & 62.44 & 52.49 & 9.95 & 78.49 & 72.54 & 5.95
            & 52.03 & 24.63 & 27.40 & 68.21 & 53.16 & 15.04 \\
DualPrompt~\cite{wang2022dualprompt}
            & 58.59 & 45.92 & 12.67 & 78.88 & 74.65 & 4.23
            & 52.07 & 23.13 & 28.94 & 72.07 & 54.93 & 17.14 \\
CODA-Prompt~\cite{smith2023coda}
            & 61.14 & 51.69 & 9.45 & 77.51 & 69.49 & 8.02
            & 58.43 & 36.51 & 21.92 & 72.63 & 55.20 & 17.44 \\
SLCA~\cite{zhang2023slca}
            & 85.05 & 72.96 & 12.09 & 92.13 & 84.41 & 7.72
            & 80.88 & 58.41 & 22.47 & 89.81 & 77.23 & 12.59 \\
SimpleCIL~\cite{zhou2025revisiting}
            & 88.82 & 72.37 & 16.45 & 94.48 & 80.29 & 14.19
            & 83.14 & 58.87 & 24.27 & 90.93 & 75.16 & 15.77 \\
APER~\cite{zhou2025revisiting}
            & 70.38 & 35.79 & 34.59 & 79.09 & 53.19 & 25.90
            & 55.31 & 43.40 & 11.91 & 72.35 & 64.29 & 8.07 \\
EASE~\cite{zhou2024expandable}
            & 65.68 & 52.88 & 12.80 & 75.07 & 70.65 & 4.42
            & 49.39 & 31.97 & 17.42 & 65.64 & 55.05 & 10.59 \\
RanPAC~\cite{mcdonnell2023ranpac}
            & 89.30 & 80.52 & 8.78 & 94.07 & 88.70 & 5.37
            & \underline{90.34} & \underline{81.32} & \underline{9.02} & 93.93 & 85.37 & 8.56 \\
MOS~\cite{sun2025mos}
            & \underline{91.07} & \underline{81.50} & 9.57 & \underline{95.38} & 91.06 & 4.32
            & 88.01 & 76.51 & 11.50 & \textbf{94.38} & \underline{87.00} & \underline{7.38} \\
TUNA~\cite{wang2025integrating}
            & 88.82 & 80.52 & \underline{8.30} & 94.46 & \underline{91.69} & \underline{2.77}
            & 87.24 & 66.10 & 21.14 & 93.41 & 79.74 & 13.67 \\
\midrule
PolyMem (Ours)
            & \textbf{92.50} & \textbf{87.90} & \textbf{4.60} & \textbf{95.90} & \textbf{93.41} & \textbf{2.48}
            & \textbf{90.67} & \textbf{84.35} & \textbf{6.32} & \underline{94.29} & \textbf{88.56} & \textbf{5.73} \\

\specialrule{0.16em}{0.25em}{0.15em}
\multirow{3}{*}{Method}
  & \multicolumn{6}{c}{ShapeNet (\textcolor{cadcolor}{$D_1$}) + OmniObject3D (\textcolor{realcolor}{$D_2$})}
  & \multicolumn{6}{c}{ShapeNet (\textcolor{cadcolor}{$D_1$}) + ShapeNet-C (\textcolor{realcolor}{$D_2$})} \\
\cmidrule(lr){2-7} \cmidrule(lr){8-13}
  & \multicolumn{3}{c}{Last Acc.}
  & \multicolumn{3}{c}{Cum. Acc.}
  & \multicolumn{3}{c}{Last Acc.}
  & \multicolumn{3}{c}{Cum. Acc.} \\
\cmidrule(lr){2-4} \cmidrule(lr){5-7} \cmidrule(lr){8-10} \cmidrule(lr){11-13}
  & \textcolor{cadcolor}{$D_1$}$\!\uparrow$ & \textcolor{realcolor}{$D_2$}$\!\uparrow$ & $\mathrm{Gap}\!\downarrow$
  & \textcolor{cadcolor}{$D_1$}$\!\uparrow$ & \textcolor{realcolor}{$D_2$}$\!\uparrow$ & $\mathrm{Gap}\!\downarrow$
  & \textcolor{cadcolor}{$D_1$}$\!\uparrow$ & \textcolor{realcolor}{$D_2$}$\!\uparrow$ & $\mathrm{Gap}\!\downarrow$
  & \textcolor{cadcolor}{$D_1$}$\!\uparrow$ & \textcolor{realcolor}{$D_2$}$\!\uparrow$ & $\mathrm{Gap}\!\downarrow$ \\
\midrule
Joint-Training
            & $\overline{88.47}$ & $\overline{86.11}$ & $\overline{2.36}$ & - & - & -
            & $\overline{87.42}$ & $\overline{85.01}$ & $\overline{2.41}$ & - & - & - \\
L2P~\cite{wang2022learning}
            & 57.08 & 34.03 & 23.05 & 75.58 & 50.31 & 25.27
            & 61.96 & 50.93 & 11.03 & 80.22 & 72.06 & 8.16 \\
DualPrompt~\cite{wang2022dualprompt}
            & 59.61 & 35.42 & 24.19 & 79.54 & 51.67 & 27.87
            & 62.69 & 52.37 & 10.32 & 77.15 & 68.33 & 8.82 \\
CODA-Prompt~\cite{smith2023coda}
            & 65.96 & 44.44 & 21.52 & 82.71 & 61.87 & 20.84
            & 69.28 & 56.98 & 12.30 & 84.06 & 74.72 & 9.34 \\
SLCA~\cite{zhang2023slca}
            & 79.03 & 59.03 & 20.00 & 91.19 & 71.64 & 19.55
            & 78.20 & 63.35 & 14.85 & 91.18 & 82.00 & 9.18 \\
SimpleCIL~\cite{zhou2025revisiting}
            & 77.52 & 53.47 & 24.05 & 88.98 & 63.59 & 25.39
            & 76.60 & 69.34 & 7.26 & 88.50 & 82.76 & 5.74 \\
APER~\cite{zhou2025revisiting}
            & 68.52 & 29.86 & 38.66 & 80.28 & 38.34 & 41.94
            & 66.75 & 59.29 & 7.46 & 79.11 & 72.31 & 6.80 \\
EASE~\cite{zhou2024expandable}
            & 58.74 & 44.44 & 14.30 & 75.29 & 61.92 & 13.37
            & 68.82 & 52.07 & 16.75 & 84.79 & 69.73 & 15.06 \\
RanPAC~\cite{mcdonnell2023ranpac}
            & 83.26 & \underline{74.69} & \underline{8.57} & 89.98 & \underline{85.10} & \underline{4.88}
            & 76.33 & 69.70 & \underline{6.62} & 90.06 & 84.34 & 5.72 \\
MOS~\cite{sun2025mos}
            & 81.95 & 70.14 & 11.81 & 92.29 & 78.01 & 14.28
            & 82.16 & 72.29 & 9.87 & 92.40 & 86.53 & 5.87 \\
TUNA~\cite{wang2025integrating}
            & \underline{84.48} & 65.28 & 19.20 & \textbf{93.20} & 76.53 & 16.67
            & \underline{83.13} & \underline{75.36} & 7.77 & \underline{92.81} & \underline{88.16} & \underline{4.65} \\
\midrule
PolyMem (Ours)
            & \textbf{87.15} & \textbf{83.52} & \textbf{3.63}
            & \underline{92.42} & \textbf{89.63} & \textbf{2.79}
            & \textbf{83.72} & \textbf{77.85} & \textbf{5.87} & \textbf{93.21} & \textbf{89.45} & \textbf{3.76} \\
\bottomrule
\end{tabular}
}
\vspace{-4mm}
\end{table*}

\begin{figure}[ht]
    \centering
    \begin{subfigure}[b]{0.49\linewidth}
        \centering
        \raisebox{0cm}{\includegraphics[width=\linewidth]{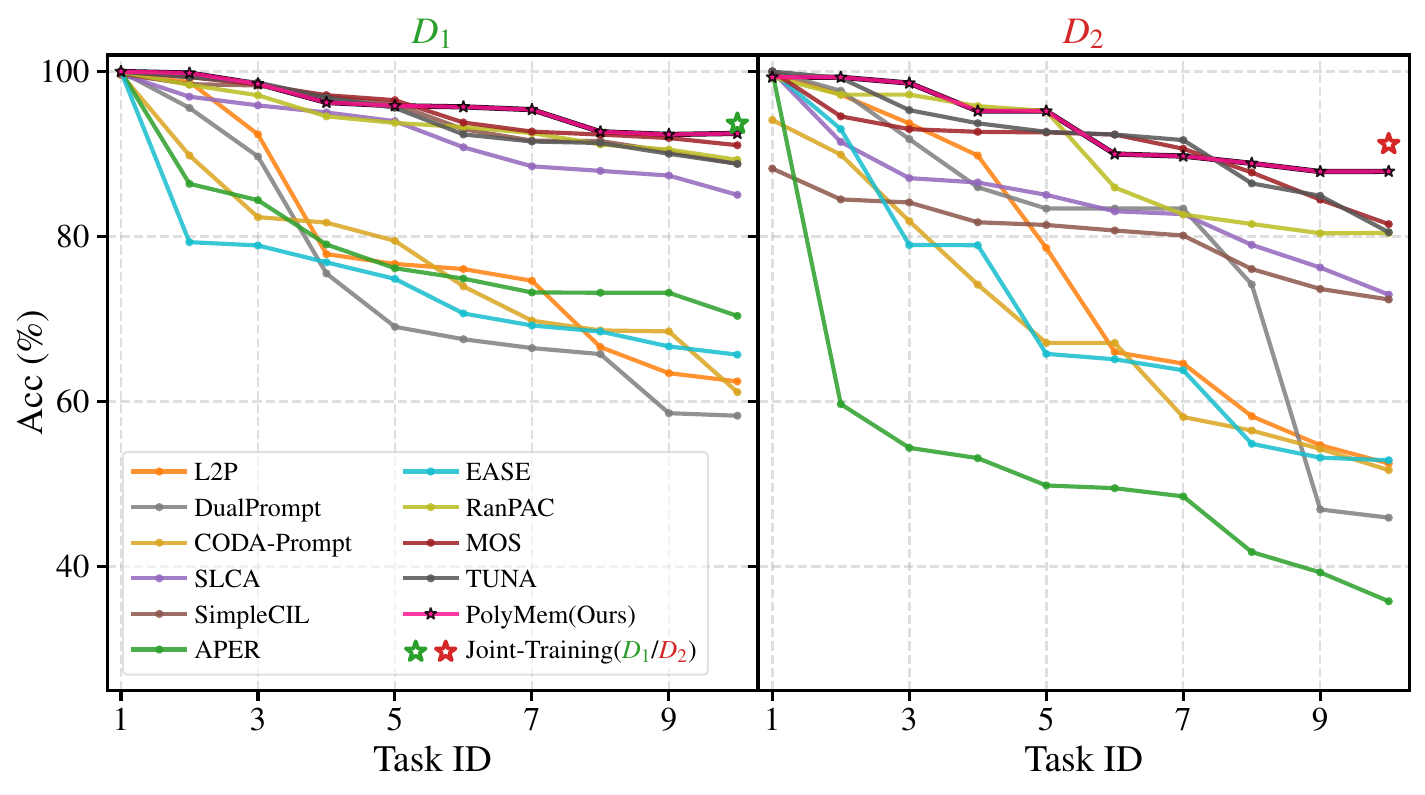}}
        \caption{ModelNet (\textcolor{cadcolor}{$D_1$}) + ScanObjectNN (\textcolor{realcolor}{$D_2$})}
        \label{fig:fig4_a}
    \end{subfigure}%
    \hspace{0.0\linewidth}%
    \begin{subfigure}[b]{0.49\linewidth}
        \centering
        \raisebox{0cm}{\includegraphics[width=\linewidth]{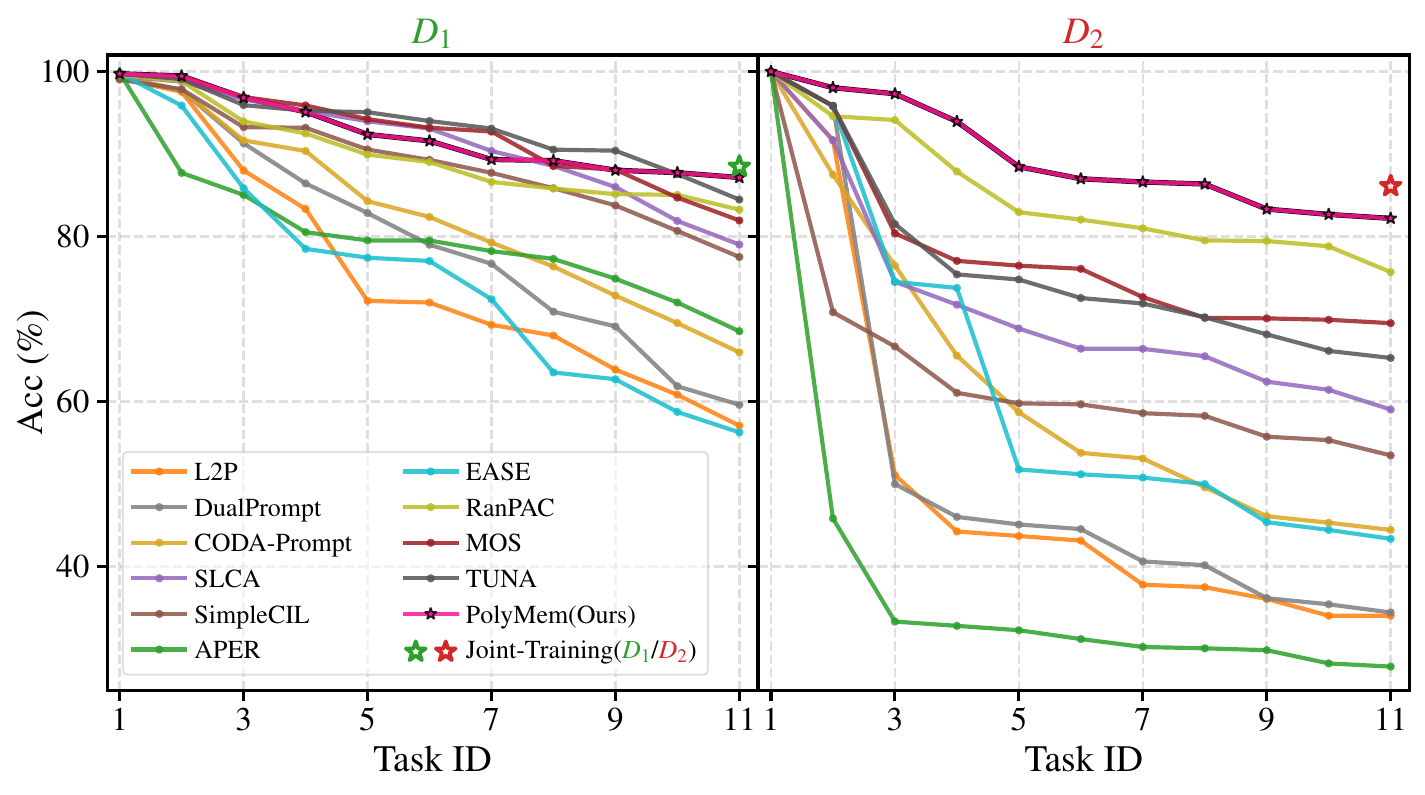}}
        \caption{ShapeNet (\textcolor{cadcolor}{$D_1$}) + OmniObject3D (\textcolor{realcolor}{$D_2$})}
        \label{fig:fig4_b}
    \end{subfigure}
    \caption{Task-wise forgetting curves showing performance discrepancy between $D_1$ and $D_2$.}
    \label{fig:fig4}
    \vspace{-4mm}
\end{figure}

As shown in Tab.~\ref{tab:ptm_results_full}, we report the results on Domain3D-CIL. The first row shows joint training (\textbf{JT}). The datasets for evaluating JT are the same as those used to compute the LA of each baseline. Thus, JT serves as the performance upper bound without catastrophic forgetting. Across different mixed datasets, the static accuracy gap between $D_1$ and $D_2$ is only around 2--3\%. For the other methods, the upper part of Tab.~\ref{tab:ptm_results_full} uses the 10-task CIL setting of \textbf{B4Inc4}, i.e., initial classes $=4$ and increment $=4$. The lower part uses the 11-task setting of \textbf{B5Inc5}. After introducing CIL, as shown in the other rows of Tab.~\ref{tab:ptm_results_full}, the performance drop on $D_2$ is more severe than that on $D_1$ compared with JT. This is reflected by the larger LA gap of each method in Tab.~\ref{tab:ptm_results_full}, compared with JT. This verifies the prevalence of performance discrepancy. The dynamic trend of performance discrepancy is clearer in Fig.~\ref{fig:fig4}.Fig.~\ref{fig:fig4_a} and Fig.~\ref{fig:fig4_b} show the task-wise forgetting curves on two mixed datasets, respectively. The vertical axis denotes the accuracy on all seen classes after each task. Comparing the two forgetting curves of the same method across domains, we observe that the curve on $D_2$ drops faster. The gap between $D_1$ and $D_2$ also tends to increase with the task ID. Comparing different methods on the same domain, all methods perform better and show smaller variance on $D_1$. In contrast, on $D_2$, different methods show clear performance differences. This demonstrates that continual learning on real or corrupted point clouds is a more challenging task. Next, we analyze the effectiveness of PolyMem. In Tab.~\ref{tab:ptm_results_full}, the best result in each column is highlighted in \textbf{bold}, and the second-best result is \underline{underlined}. PolyMem achieves the lowest LA and CA gaps across all mixed datasets, effectively mitigating performance discrepancy. Compared with any baseline, PolyMem achieves an average improvement of at least 6.8\% on $D_2$ in Tab.~\ref{tab:ptm_results_full}. This improvement is not obtained at the cost of performance on $D_1$. Instead, PolyMem slightly improves or maintains a $D_1$ performance close to JT. Fig.~\ref{fig:fig4} further shows that the anti-forgetting advantage of PolyMem continues to increase in later tasks.

\subsection{Ablation studies}
\begin{figure}[h]
    \centering
    \captionsetup{skip=2pt}

    \begin{minipage}[t]{0.49\textwidth}
        \centering
        \begin{subfigure}[t]{0.49\linewidth}
            \centering
            \raisebox{0.15em}{%
                \includegraphics[height=0.14\textheight,keepaspectratio]{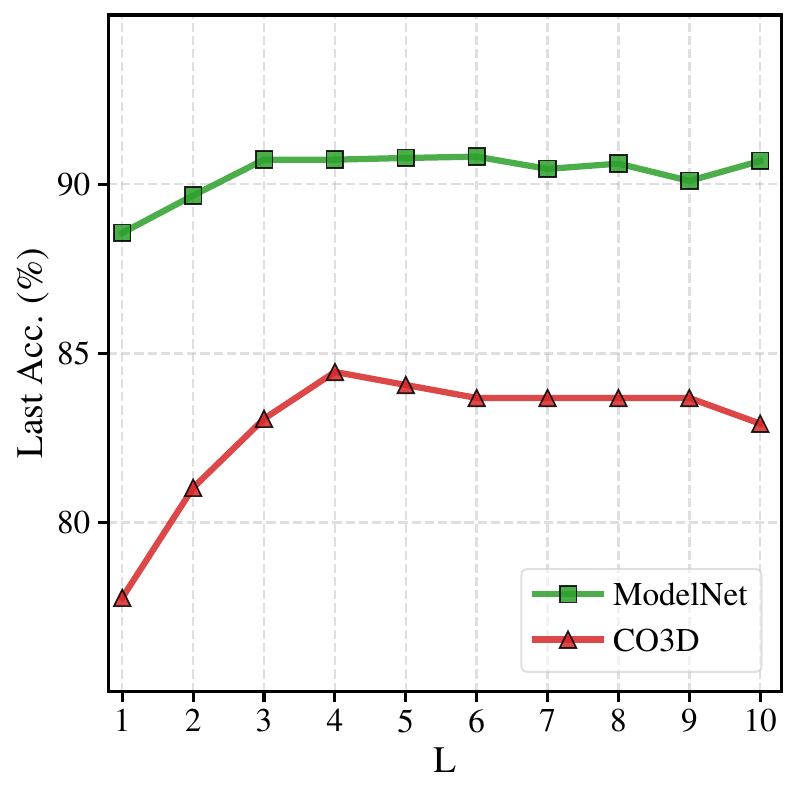}%
            }
            \label{fig:fig7_a}
        \end{subfigure}
        \hfill
        \begin{subfigure}[t]{0.49\linewidth}
            \centering
            \raisebox{0.12em}{%
                \includegraphics[height=0.142\textheight,keepaspectratio]{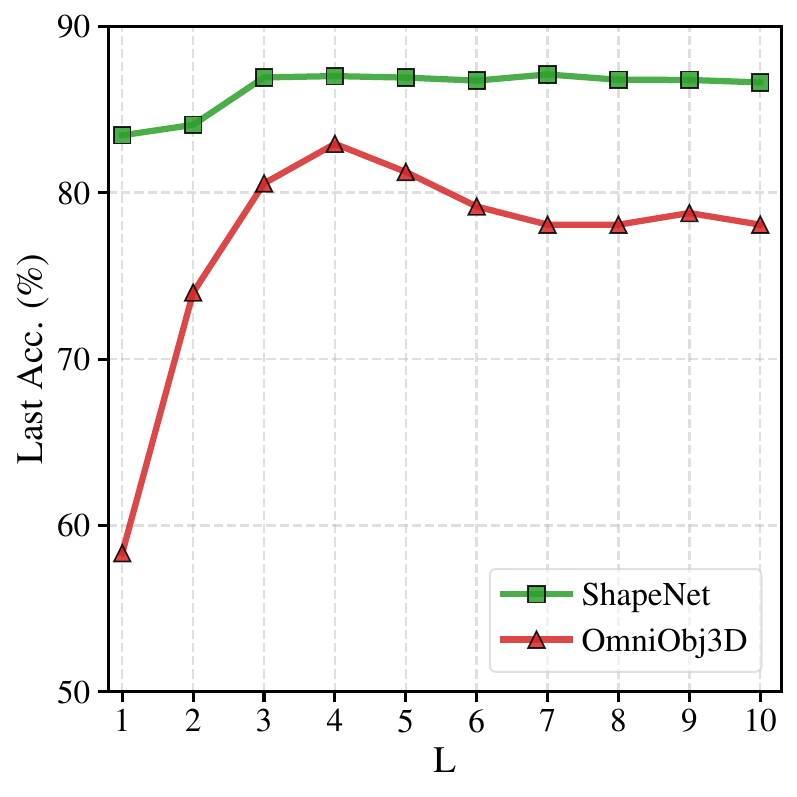}%
            }
            \label{fig:fig7_b}
        \end{subfigure}

        \caption{Ablation on $L$}
        \label{fig:fig7}
    \end{minipage}
    \hfill
    \begin{minipage}[t]{0.497\textwidth}
        \centering

        \begin{minipage}[t]{0.495\linewidth}
            \centering
            \includegraphics[height=0.142\textheight,keepaspectratio]{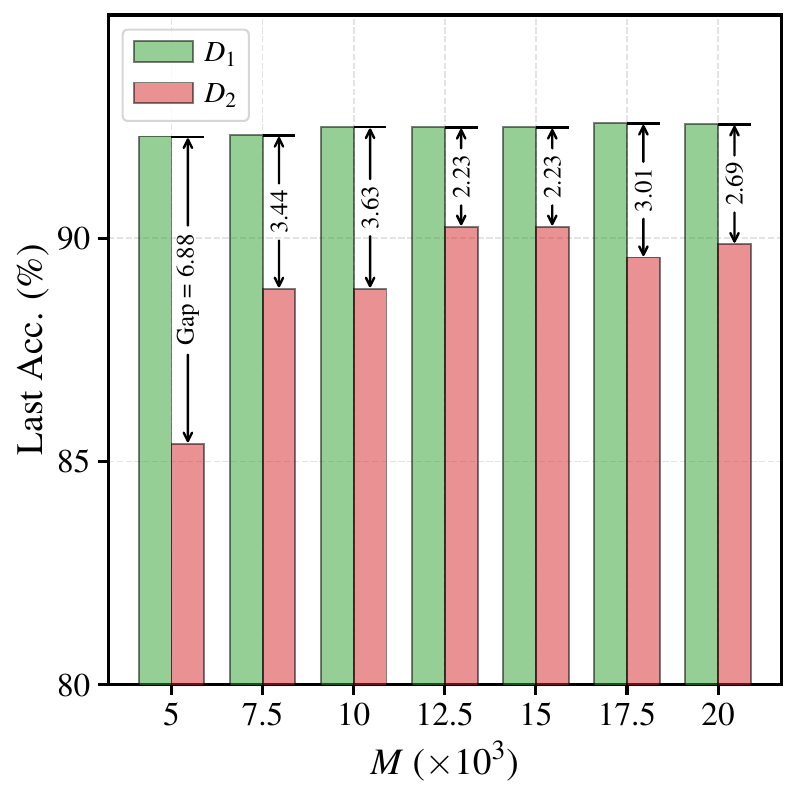}
            \caption{Ablation on $M$}
            \label{fig:fig8}
        \end{minipage}
        \hfill
        \begin{minipage}[t]{0.49\linewidth}
            \centering
            \includegraphics[height=0.142\textheight,keepaspectratio]{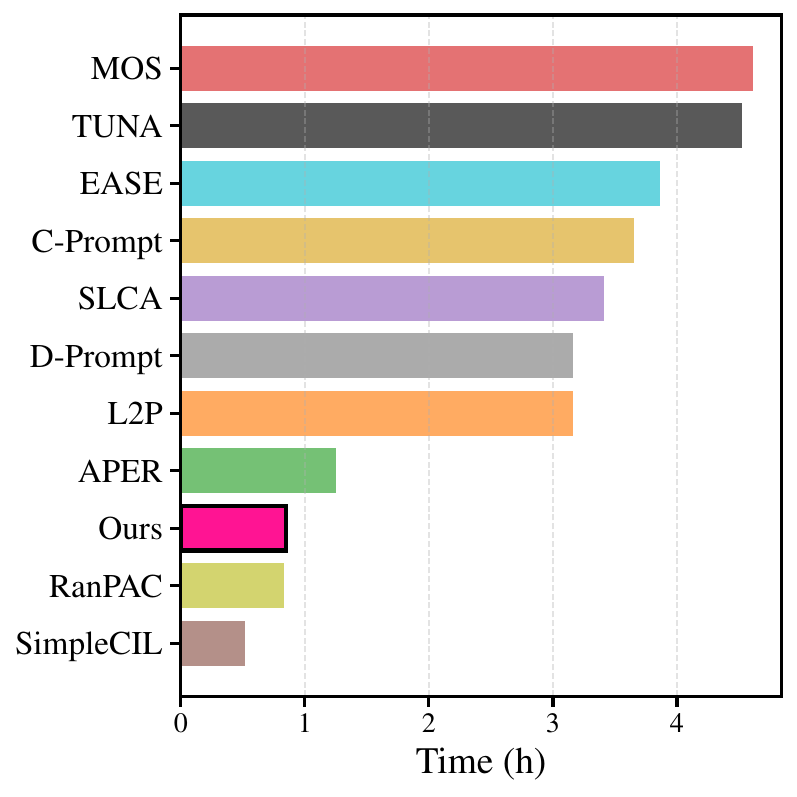}
            \caption{Running Time}
            \label{fig:fig9}
        \end{minipage}

    \end{minipage}
\vspace{-2mm}
\end{figure}

\label{sec:ablation}
In Sec.~\ref{sec:main_exp}, the hyperparameters of PolyMem are fixed to $L=4$ and $M=10000$. $M$ is set the same as in RanPAC for a fair comparison. In this section, we further analyze the effect of the feature-interaction order $L$ and the stability of PolyMem to the projected-space dimension $M$. Fig.~\ref{fig:fig7} analyzes the effect of $L$ on two mixed datasets. When $L$ increases from 1 to 4, the performance improves clearly on both $D_1$ and $D_2$, with a smaller Last Acc. gap between the two domains. This suggests that the introduced higher-order interactions can better characterize the complex feature distributions of real or reconstructed point clouds, and even CAD point clouds. However, further increasing $L$ does not lead to continuous gains. The performance on $D_1$ gradually saturates, while that on $D_2$ slightly decreases. This is because, under a fixed $M$, a larger $L$ introduces higher-order feature interactions, but also reduces the sketch dimension $M_n$ allocated to each order. As a result, the approximation quality of existing low- and middle-order interactions may decrease. Therefore, an overly large $L$ cannot improve the discriminative ability without bound. Overall, $L=4$ achieves a good balance between interaction modeling capacity and sketch approximation quality. This indicates that the gain of PolyMem mainly comes from moderate higher-order feature-interaction modeling, rather than simply using higher and higher orders. As shown in Fig.~\ref{fig:fig8}, the ablation on $M$ shows that PolyMem remains stable once the $M$ reaches a moderate scale. When $M$ is small, e.g., $M=5000$, the overly limited sketch dimension leads to a large Last Acc. gap between $D_1$ and $D_2$. Under a fixed order $L$, a smaller $M$ also reduces the $M_n$, increasing the approximation error of polynomial feature interactions at each order. Real point clouds are more sensitive to high-order feature-interaction modeling and require sufficient $M_n$. As $M$ increases from a small value to a moderate scale, the performance on $D_2$ improves and the domain gap clearly decreases. Further increasing $M$ yields stable performance, indicating that PolyMem does not rely on an overly large $M$. When $M=10000$, RanPAC and PolyMem store 110M and 100M parameters respectively. In contrast, directly storing explicit 3rd- or 4th-order feature interactions would require about 1B and 1T parameters. This highlights the benefit: PolyMem captures high-order interactions implicitly, instead of explicitly storing high-order tensors. Fig.~\ref{fig:fig9} shows the time efficiency of different methods. The total training and testing time of PolyMem is much lower than that of most baselines, and is comparable to RanPAC.

We provide implementation and dataset details, additional task-split experiments, results with different backbones, classic CIL baseline results in Fig.~\ref{fig:fig1_b}, and more experimental details in the \textbf{Appendix}.

\section{Conclusion.}
We formalize and study performance discrepancy in cross-domain 3D CIL, and introduce the Domain3D-CIL protocol to systematically evaluate this problem on heterogeneous point cloud.
Experiments show that performance discrepancy is prevalent across CIL baselines.
To address this issue, we propose PolyMem, which implicitly captures high-order interactions in heterogeneous 3D feature distributions.
Experiments show that PolyMem improves performance on both domains while effectively reducing the cross-domain performance gap.
However, this paper does not analyze all causes of performance discrepancy across all baselines.
Future work will further investigate the mechanisms behind performance discrepancy in different methods.

\clearpage
{
    \small
    \bibliographystyle{plain}
    \bibliography{neurips_2026}
}

\clearpage
\section{Appendix}
\subsection{Theoretical Background}
\label{app:theoretical_background}
This subsection provides background knowledge on distributional statistics,
polynomial feature interactions, kernel approximation, and closed-form ridge
regression.  We emphasize that the materials in this subsection are standard
concepts and tools, and are \textbf{NOT methodological contributions} of this work.
\subsubsection{Distributional Statistics}
\label{app:statistics_bg}

A statistic is a quantity computed from data to summarize a property of an
underlying distribution.
Moments are a fundamental class of statistics that describe the shape of a
distribution through expectations of powers of a random variable.
For a scalar random variable $X$, the $r$-th raw moment and the $r$-th central
moment are defined as
\begin{equation}
    m_r=\mathbb{E}[X^r],
    \qquad
    \mu_r=\mathbb{E}\big[(X-\mu)^r\big],
\end{equation}
where $\mu=\mathbb{E}[X]$ is the mean.
The mean and variance are the most common low-order summaries:
\begin{equation}
    \mu=\mathbb{E}[X],
    \qquad
    \sigma^2=\mathbb{E}\big[(X-\mu)^2\big].
\end{equation}
The mean describes the central location of a distribution, while the variance
describes its dispersion.

Beyond the first two orders, higher-order moments characterize more detailed
properties of the distributional shape.
The third central moment measures the asymmetry of a distribution around its
mean.
Its standardized form is known as skewness~\cite{pearson1895skewness}:
\begin{equation}
    \gamma_1
    =
    \frac{\mathbb{E}\big[(X-\mu)^3\big]}{\sigma^3}.
\end{equation}
A positive skewness indicates that the distribution has a longer or heavier
right tail, while a negative skewness indicates a longer or heavier left tail.
When the distribution is symmetric around its mean, the third central moment is
zero.
Thus, skewness captures information that cannot be reflected by the mean or
variance alone.

The fourth central moment characterizes the concentration and tail behavior of
a distribution.
Its standardized form is called kurtosis~\cite{pearson1905rejoinder}:
\begin{equation}
    \gamma_2
    =
    \frac{\mathbb{E}\big[(X-\mu)^4\big]}{\sigma^4}.
\end{equation}
A distribution with larger kurtosis typically places more probability mass in
the tails and around the center, compared with a distribution with lighter
tails.
Kurtosis is therefore commonly used to describe heavy-tailed behavior and the
frequency of extreme deviations from the mean.

\subsubsection{Polynomial Feature Interactions}
\label{app:poly_interactions_bg}

The previous subsection introduces moments as statistics obtained from
expectations of powers of random variables.
Polynomial feature interactions provide the corresponding feature-level
building blocks for such statistics.
The key distinction is that a polynomial interaction from a single sample is
not yet a statistic; it becomes a statistic only after taking expectation over
a distribution or empirical aggregation over samples.

To see this in the simplest case, consider a scalar random variable $X$.
The polynomial terms $X^2$, $X^3$, and $X^4$ become raw moments after taking
expectation:
\begin{equation}
    \mathbb{E}[X^2],
    \qquad
    \mathbb{E}[X^3],
    \qquad
    \mathbb{E}[X^4].
\end{equation}
Central moments can be computed from these raw moments.
For example, the second central moment is
\begin{equation}
    \mathbb{E}[(X-\mu)^2]
    =
    \mathbb{E}[X^2]-\mu^2,
\end{equation}
and the third central moment is
\begin{equation}
    \mathbb{E}[(X-\mu)^3]
    =
    \mathbb{E}[X^3]
    -3\mu\mathbb{E}[X^2]
    +2\mu^3.
\end{equation}
Thus, higher-order statistics such as variance and skewness are formed from
aggregated polynomial terms, rather than from isolated single-sample products.

For a vector $F\in\mathbb{R}^{D}$, the same idea extends to cross-dimensional
statistics.
An $n$-th order polynomial interaction is a degree-$n$ product of feature
dimensions:
\begin{equation}
    F_{d_1}F_{d_2}\cdots F_{d_n}.
\end{equation}
For a single sample, this is only a feature-level product.
After aggregation, it forms a high-order raw moment or cross-moment:
\begin{equation}
    \mathbb{E}\!\left[
    F_{d_1}F_{d_2}\cdots F_{d_n}
    \right],
    \qquad
    \frac{1}{N}\sum_{i=1}^{N}
    F_{i,d_1}F_{i,d_2}\cdots F_{i,d_n}.
\end{equation}
For example, second-order products $F_pF_q$ give
\begin{equation}
    \frac{1}{N}\sum_{i=1}^{N}F_{i,p}F_{i,q},
\end{equation}
which is a second-order raw cross-moment.
When $p=q$, it reduces to a raw second moment of one feature dimension; after
centering, the diagonal case corresponds to variance.
Similarly, third-order products $F_pF_qF_r$ give
\begin{equation}
    \frac{1}{N}\sum_{i=1}^{N}F_{i,p}F_{i,q}F_{i,r},
\end{equation}
which is a third-order raw moment or cross-moment.
After centering, the one-dimensional case $p=q=r$ corresponds to the numerator
of skewness.
Therefore, polynomial feature interactions are elementary components from
which high-order distributional statistics can be constructed through
sample-level aggregation.

Polynomial interactions also arise naturally from polynomial kernels.
Given two vectors $F_i,F_j\in\mathbb{R}^{D}$, the degree-$n$ polynomial inner
product
\begin{equation}
    (F_i^\top F_j)^n
\end{equation}
can be viewed as an inner product in a degree-$n$ ordered monomial feature
space, or equivalently a tensor-product feature space.
For instance, the quadratic case expands as
\begin{equation}
    (F_i^\top F_j)^2
    =
    \left(\sum_{d=1}^{D}F_{i,d}F_{j,d}\right)^2
    =
    \sum_{p=1}^{D}\sum_{q=1}^{D}
    F_{i,p}F_{i,q}F_{j,p}F_{j,q}.
\end{equation}
This expansion shows that the quadratic kernel compares all ordered
second-degree feature products between $F_i$ and $F_j$.
Higher-degree polynomial kernels analogously compare higher-degree ordered
feature products.

However, explicitly storing all polynomial interaction terms is impractical.
A full ordered $n$-th order interaction tensor over a $D$-dimensional vector
contains $D^n$ entries.
For $D=1024$, this gives
\begin{equation}
    D^2 = 1024^2 = 1{,}048{,}576,
\end{equation}
\begin{equation}
    D^3 = 1024^3 = 1{,}073{,}741{,}824,
\end{equation}
and
\begin{equation}
    D^4 = 1024^4 = 1{,}099{,}511{,}627{,}776
\end{equation}
entries for second-, third-, and fourth-order interactions, respectively.
Using 32-bit floating point storage, these correspond to approximately
$4$ MiB, $4$ GiB, and $4$ TiB for a single full tensor.
The cost would further increase if such tensors were maintained separately for
multiple classes.

This distinction is important for PolyMem.
PolyMem does not explicitly recover or store each high-order moment tensor.
Instead, it constructs a compact polynomially lifted feature space and
accumulates ridge-regression statistics in that space.
In this sense, polynomial feature interactions make high-order feature products
available for statistical accumulation, while the compact approximation
introduced in the next subsection keeps the representation memory-efficient.

\subsubsection{RBF Kernel}
\label{app:kernel_taylor_bg}

A kernel function provides a way to measure the similarity between two inputs.
From the perspective of kernel methods, a positive semi-definite kernel
$k(\cdot,\cdot)$ corresponds to an implicit feature mapping $\phi(\cdot)$ such
that
\begin{equation}
    k(F_i,F_j)=\phi(F_i)^\top\phi(F_j).
\end{equation}
Therefore, a nonlinear similarity function in the original space can be
interpreted as a linear inner product in a mapped feature space.
This viewpoint is useful when the mapped feature space is high-dimensional or
even infinite-dimensional, because the kernel value can be evaluated without
explicitly constructing all mapped features.

The radial basis function (RBF) kernel is one of the most widely used kernels.
For two vectors $F_i,F_j\in\mathbb{R}^{D}$, it is defined as
\begin{equation}
    k_{\mathrm{RBF}}(F_i,F_j)
    =
    \exp\!\left(-\gamma\|F_i-F_j\|_2^2\right),
\end{equation}
where $\gamma>0$ controls the bandwidth of the kernel.
A larger $\gamma$ makes the similarity decay faster as the distance between
$F_i$ and $F_j$ increases, while a smaller $\gamma$ gives a smoother similarity
measure.

In PolyMem, the backbone features are L2-normalized before constructing the
polynomial mapping.
For normalized features satisfying $\|F_i\|_2=\|F_j\|_2=1$, the squared
Euclidean distance can be written as
\begin{equation}
    \|F_i-F_j\|_2^2
    =
    \|F_i\|_2^2+\|F_j\|_2^2-2F_i^\top F_j
    =
    2-2F_i^\top F_j.
\end{equation}
Substituting this into the RBF kernel gives
\begin{equation}
\begin{aligned}
    k_{\mathrm{RBF}}(F_i,F_j)
    &=
    \exp\!\left(-\gamma(2-2F_i^\top F_j)\right) \\
    &=
    \exp(-2\gamma)\exp(2\gamma F_i^\top F_j).
\end{aligned}
\end{equation}

The second exponential term can be expanded using the Taylor series:
\begin{equation}
    \exp(2\gamma F_i^\top F_j)
    =
    \sum_{n=0}^{\infty}
    \frac{(2\gamma)^n}{n!}
    (F_i^\top F_j)^n.
\end{equation}
Thus, the RBF kernel can be decomposed as
\begin{equation}
    k_{\mathrm{RBF}}(F_i,F_j)
    =
    \exp(-2\gamma)
    \sum_{n=0}^{\infty}
    \frac{(2\gamma)^n}{n!}
    (F_i^\top F_j)^n.
\end{equation}

This decomposition connects the RBF kernel to the polynomial feature
interactions introduced in Appendix~\ref{app:poly_interactions_bg}.
The zeroth-order term is a constant term, the first-order term corresponds to
linear interactions, the second-order term corresponds to quadratic
interactions, and higher-order terms correspond to higher-degree polynomial
interactions.
Therefore, the RBF kernel provides an order-wise decomposition that contains
polynomial interactions of all degrees.

PolyMem uses a finite-order approximation by retaining the terms up to order
$L$:
\begin{equation}
    k_{\mathrm{Poly}}^{(L)}(F_i,F_j)
    =
    \exp(-2\gamma)
    \sum_{n=0}^{L}
    \frac{(2\gamma)^n}{n!}
    (F_i^\top F_j)^n.
\end{equation}
The order $L$ controls the maximum degree of polynomial interaction included in
the mapped feature space.
In this way, the RBF Taylor expansion provides a structured and order-wise
basis for the PolyMem mapping, while the next subsection explains how these
polynomial terms can be compactly approximated by TensorSketch.

\subsubsection{Approximation by TensorSketch}
\label{app:tensorsketch_bg}

The RBF Taylor expansion in Appendix~\ref{app:kernel_taylor_bg} decomposes the
RBF kernel into polynomial interactions of different orders.
However, directly constructing the explicit monomial feature map for each order
is impractical.
For the $n$-th order term, the explicit feature map contains degree-$n$
products of the input dimensions, whose size grows on the order of
$\mathcal{O}(D^n)$.
As discussed in Appendix~\ref{app:poly_interactions_bg}, this cost becomes
prohibitive even for moderate feature dimensions.

TensorSketch~\cite{pham2013fast} provides a compact randomized approximation
to such polynomial feature maps.
Let $\psi_n(F)$ denote the explicit degree-$n$ monomial feature map.
It satisfies
\begin{equation}
    \psi_n(F_i)^\top \psi_n(F_j)
    =
    (F_i^\top F_j)^n .
\end{equation}
Instead of materializing $\psi_n(F)$, TensorSketch constructs a compact feature
\begin{equation}
    \mathrm{TS}_n(F)\in\mathbb{R}^{M_n},
    \qquad M_n \ll D^n,
\end{equation}
such that
\begin{equation}
    \mathrm{TS}_n(F_i)^\top \mathrm{TS}_n(F_j)
    \approx
    (F_i^\top F_j)^n .
\end{equation}
Therefore, TensorSketch keeps the polynomial-kernel inner product approximately
available while avoiding the explicit $\mathcal{O}(D^n)$ expansion.

PolyMem applies TensorSketch separately to each polynomial order in the
finite-order RBF Taylor expansion.
For the $n$-th order term, PolyMem uses a sketch feature
$\mathrm{TS}_n(F)\in\mathbb{R}^{M_n}$ and scales it according to the
corresponding Taylor coefficient.
Specifically, the order-$n$ block is written as
\begin{equation}
    e^{-\gamma}
    \sqrt{\frac{(2\gamma)^n}{n!}}
    \mathrm{TS}_n(F).
\end{equation}
The square-root scaling is used because the kernel approximation is obtained by
taking an inner product between two mapped features.
When two order-$n$ blocks are multiplied, the product of the two scaling factors
recovers the coefficient
$e^{-2\gamma}(2\gamma)^n/n!$ in the RBF Taylor expansion.

By concatenating the constant term and all order-wise sketch features, PolyMem
defines
\begin{equation}
\phi_{\mathrm{Poly}}(F)
=
\left[
e^{-\gamma},
\;
e^{-\gamma}\sqrt{\frac{2\gamma}{1!}}\mathrm{TS}_1(F),
\;
\dots,
\;
e^{-\gamma}\sqrt{\frac{(2\gamma)^L}{L!}}\mathrm{TS}_L(F)
\right]
\in\mathbb{R}^{M},
\end{equation}
where the sketch dimensions satisfy
\begin{equation}
    1+\sum_{n=1}^{L}M_n=M.
\end{equation}
With this construction, the inner product between two PolyMem features
approximates the finite-order RBF Taylor kernel:
\begin{equation}
\begin{aligned}
    \phi_{\mathrm{Poly}}(F_i)^\top \phi_{\mathrm{Poly}}(F_j)
    &\approx
    e^{-2\gamma}
    \sum_{n=0}^{L}
    \frac{(2\gamma)^n}{n!}
    (F_i^\top F_j)^n  \\
    &=
    k_{\mathrm{Poly}}^{(L)}(F_i,F_j).
\end{aligned}
\end{equation}
Thus, TensorSketch enables PolyMem to retain structured order-wise polynomial
interactions in a fixed-dimensional mapped feature space, without explicitly
storing the full high-order interaction tensors.

\subsubsection{Ridge Regression}
\label{app:ridge_bg}

The previous subsections explain how PolyMem constructs a compact mapped
feature $F_i'=\phi_{\mathrm{Poly}}(F_i)\in\mathbb{R}^{M}$ that approximates the
finite-order RBF Taylor feature space.
After this mapping is fixed, the remaining question is how to learn a
classifier from these mapped features without storing previous-task samples.
PolyMem follows the closed-form ridge regression formulation~\cite{hoerl1970ridge}, which naturally
supports additive statistic accumulation.

For each sample $x_i$, let $F_i'=\phi_{\mathrm{Poly}}(F_i)$ denote its mapped
feature, and let $e_{y_i}\in\mathbb{R}^{C}$ denote the one-hot label vector.
A linear classifier in the mapped feature space predicts
\begin{equation}
    s(x_i)=W^\top F_i',
\end{equation}
where $W\in\mathbb{R}^{M\times C}$ is the classifier weight matrix.
Ridge regression learns $W$ by minimizing the regularized squared loss:
\begin{equation}
    \min_W
    \sum_{i=1}^{N}
    \left\|W^\top F_i' - e_{y_i}\right\|_2^2
    +
    \lambda \|W\|_F^2,
\end{equation}
where $\lambda$ is the ridge coefficient.
The regularization term improves numerical stability and controls the magnitude
of the classifier weights.

This objective has a closed-form solution.
Taking the derivative with respect to $W$ and setting it to zero gives the
normal equation
\begin{equation}
    (G+\lambda I)W = Q,
\end{equation}
where
\begin{equation}
    G=\sum_{i=1}^{N}F_i'{F_i'}^{\top}\in\mathbb{R}^{M\times M},
    \qquad
    Q=\sum_{i=1}^{N}F_i'e_{y_i}^{\top}\in\mathbb{R}^{M\times C}.
\end{equation}
Thus, the classifier can be obtained as
\begin{equation}
    W=(G+\lambda I)^{-1}Q.
\end{equation}
Here, $G$ stores the uncentered second-order statistic of the mapped features,
while $Q$ stores the correlation between mapped features and class labels.

The key property for class-incremental learning is that both $G$ and $Q$ are
additive over samples.
Suppose task $t$ provides the current training set $\mathcal{D}_t$.
Instead of recomputing statistics from all previous data, we only need to add
the current-task contributions:
\begin{equation}
    G \leftarrow G+
    \sum_{(x_i,y_i)\in\mathcal{D}_t}
    F_i'{F_i'}^{\top},
    \qquad
    Q \leftarrow Q+
    \sum_{(x_i,y_i)\in\mathcal{D}_t}
    F_i'e_{y_i}^{\top}.
\end{equation}
After the update, solving $(G+\lambda I)W=Q$ gives the classifier over all
classes represented in the accumulated statistics.
Therefore, previous-task samples do not need to be stored.

This additive formulation is valid only when all tasks are represented in a
consistent feature space.
This is why PolyMem initializes $\phi_{\mathrm{Poly}}(\cdot)$ once at the first
task and keeps it fixed afterwards.
With a fixed mapping, the statistics accumulated from different tasks are
compatible, and the memory cost is determined by the stored matrices
$G\in\mathbb{R}^{M\times M}$ and $Q\in\mathbb{R}^{M\times C}$, rather than by
the number of training samples.
This directly leads to the task-wise PolyMem procedure summarized in
Algorithm~\ref{alg:polymem}.

\subsection{Algorithmic Description of PolyMem}
\label{app:polymem_algorithm}

We summarize the task-wise training procedure of PolyMem in
Algorithm~\ref{alg:polymem}.
The construction of the PolyMem mapping
$\phi_{\mathrm{Poly}}(\cdot)$ follows the finite-order RBF Taylor expansion
and the TensorSketch-based polynomial approximation introduced in
Appendix~\ref{app:kernel_taylor_bg} and
Appendix~\ref{app:tensorsketch_bg}.
Here, we focus on how this fixed mapping is used for statistic accumulation and
closed-form classifier estimation in the PTM-based CIL setting.

Given an input sample $x_i$, the pre-trained backbone extracts a feature
$F_i=f(x_i)\in\mathbb{R}^{D}$, which is mapped to
$F_i'=\phi_{\mathrm{Poly}}(F_i)\in\mathbb{R}^{M}$.
The mapping $\phi_{\mathrm{Poly}}(\cdot)$ is initialized once at the first task
and then kept fixed throughout the incremental learning process.
This fixed-mapping design ensures that the accumulated statistics from all tasks
are represented in the same polynomially lifted feature space.
At each task, PolyMem only uses the current-task training samples to update the
statistics $Q$ and $G$ defined in Eq.~\ref{Eq:ridge_regression}.
The classifier is then obtained by the closed-form ridge regression solution
described in Appendix~\ref{app:ridge_bg}.  For a fair comparison with RanPAC~\cite{mcdonnell2023ranpac}, the ridge
coefficient is selected from the same fixed candidate set
$\Lambda=\{10^{-8},10^{-7},\dots,10^{7},10^{8}\}$. Therefore, PolyMem does not store raw samples or raw backbone features from previous tasks.

\begin{algorithm}[ht]
\caption{PolyMem for PTM-based Class-Incremental Learning}
\label{alg:polymem}
\begin{algorithmic}[1]
\REQUIRE Incremental training sets $\{\mathcal{D}_t\}_{t=0}^{T-1}$; backbone $f(\cdot)$; total number of classes $C$; PolyMem dimension $M$; Taylor order $L$; ridge candidate set $\Lambda$
\ENSURE Classifier weights for all seen classes after each task

\STATE Initialize $Q\leftarrow \mathbf{0}\in\mathbb{R}^{M\times C}$,
$G\leftarrow \mathbf{0}\in\mathbb{R}^{M\times M}$, and PolyMem mapping parameters as empty

\FOR{$t=0,1,\dots,T-1$}
    \STATE Let $\mathcal{Y}_t$ be the classes introduced at task $t$, and extract
    $F_i=f(x_i)$ for all $(x_i,y_i)\in\mathcal{D}_t$

    \IF{$t=0$}
        \STATE Initialize the fixed mapping $\phi_{\mathrm{Poly}}(\cdot)$ with Taylor order $L$ and dimension $M$; set
        $\gamma=\left(2\,\mathrm{median}_{(i,j)\in\mathcal{P}_0}\|\bar{F}_i-\bar{F}_j\|_2^2\right)^{-1}$
        on randomly sampled task-0 normalized feature pairs unless $\gamma$ is manually specified
    \ENDIF

    \STATE Map current-task features as $F_i'=\phi_{\mathrm{Poly}}(F_i)\in\mathbb{R}^{M}$ and construct one-hot labels $e_{y_i}\in\mathbb{R}^{C}$

    \STATE Update accumulated statistics:
    \[
        Q \leftarrow Q+\sum_{(x_i,y_i)\in\mathcal{D}_t}F_i'e_{y_i}^{\top},
        \qquad
        G \leftarrow G+\sum_{(x_i,y_i)\in\mathcal{D}_t}F_i'{F_i'}^{\top}.
    \]

    \STATE Select $\lambda_t\in\Lambda$ using a validation split of the current task and solve
    \[
        W_t=(G+\lambda_t I)^{-1}Q .
    \]
    \STATE Use the columns of $W_t$ corresponding to seen classes
    $\bigcup_{j=0}^{t}\mathcal{Y}_j$ for classification
\ENDFOR
\end{algorithmic}
\end{algorithm}

\subsection{Dataset Details}
\label{app:datasets}

\noindent
This appendix summarizes the source datasets, mixed benchmarks, preprocessing, and semantic alignment used in Domain3D-CIL.
Unless otherwise stated, point clouds are XYZ-only, resampled to \(N{=}1024\) points per shape, centered by subtracting the point-set centroid, and isotropically scaled by the maximum absolute coordinate so that all points fit inside \([-1,1]^3\). We refer to this preprocessing as \emph{unit-cube normalization}.

\paragraph{Domain tag \texttt{type} vs.\ training.}
Each local HDF5 record stores a string field \texttt{type} indicating the data source or corruption type, e.g., \texttt{clean}, \texttt{real}, \texttt{noisy}, or a specific corruption identifier.
\textbf{The \texttt{type} field is used only for bookkeeping and domain-stratified evaluation; it is not exposed to the model during incremental training}, consistent with the main protocol where domain labels are unavailable during training.

\subsubsection{Source Datasets}
\label{app:sources}

\paragraph{ModelNet40~\cite{wu20153d}.}
ModelNet40 contains synthetic CAD objects from common indoor and household categories.
We use the standard \textbf{40}-class split with \textbf{train/test = 9{,}840/2{,}468} shapes, totaling \textbf{12{,}308} shapes.

\paragraph{ShapeNetCore(v2)~\cite{chang2015shapenet}.}
We adopt a \textbf{fixed 55-class} point-cloud taxonomy, used consistently across ShapeNet, overlapping OmniObject3D categories, and ShapeNet-C.
This taxonomy is not intended to enumerate the full ShapeNet synset inventory.
The commonly used split contains \textbf{train/validation/test = 35{,}708/5{,}158/10{,}261} shapes.
Following our experimental protocol, we use only the training and test partitions, resulting in \textbf{train/test = 35{,}708/10{,}261} shapes, totaling \textbf{45{,}969} shapes.
The validation partition is not used in our benchmark construction.

\paragraph{ScanObjectNN~\cite{uy2019revisiting}.}
ScanObjectNN contains real indoor object scans captured with RGB-D sensors.
We use the \textbf{15} categories in the official \texttt{shape\_names\_ext} order distributed with the HKUST reference materials.
The full split contains \textbf{train/test = 2{,}309/581} scans, totaling \textbf{2{,}890} scans.

\paragraph{CO3D~\cite{reizenstein2021common}.}
CO3D provides real object reconstructions from crowdsourced videos.
We use the official \textbf{sequence-level reconstructed point clouds} in PLY format bundled with CO3D v2, typically corresponding to one fused reconstruction per video sequence rather than per-frame point clouds.
For the ModelNet40 + CO3D benchmark, we retain \textbf{13} categories that can be aligned with ModelNet40.
After an approximately \textbf{80/20} stratified split per class with a fixed random seed, the retained subset contains \textbf{train/test = 8{,}823/2{,}205} point clouds, totaling \textbf{11{,}028} point clouds.

\paragraph{OmniObject3D~\cite{wu2023omniobject3d}.}
OmniObject3D contains high-quality real-scanned 3D objects over \textbf{216} categories.
Before semantic filtering and merging with ShapeNet, the split used here contains \textbf{train/test = 4{,}641/1{,}270} scans, totaling \textbf{5{,}911} scans.

\paragraph{ShapeNet-C.}
ShapeNet-C denotes corrupted ShapeNet counterparts constructed following PointCloud-C~\cite{ren2022benchmarking}.
We use the complete ShapeNet-C bank constructed from ShapeNet, covering the following corruption families:
\texttt{scale}, \texttt{jitter}, \texttt{rotate}, \texttt{dropout\_global}, \texttt{dropout\_local}, \texttt{add\_global}, and \texttt{add\_local}.
In the ShapeNet + ShapeNet-C benchmark, corrupted samples are randomly selected from this ShapeNet-C bank following the mixing protocol described in \S\ref{app:shapenet_c_mix}.

\subsubsection{Mixed Benchmarks: Filtered Cardinality}
\label{app:filtered}

After semantic alignment, the number of non-CAD samples actually merged into each benchmark is as follows:
\begin{itemize}\itemsep2pt
\item \textbf{ModelNet40 + ScanObjectNN:} total \textbf{train/test = 11{,}844/2{,}971}. ModelNet40 contributes \textbf{9{,}840/2{,}468} CAD samples, and the semantically matched ScanObjectNN subset contributes \textbf{2{,}004/503} real scans. The full ScanObjectNN split is not entirely used; only categories with a ModelNet40 mapping are retained.

\item \textbf{ModelNet40 + CO3D:} total \textbf{train/test = 18{,}663/4{,}673}. ModelNet40 contributes \textbf{9{,}840/2{,}468} CAD samples, and CO3D contributes \textbf{8{,}823/2{,}205} real reconstructed point clouds from the 13 aligned categories.

\item \textbf{ShapeNet + OmniObject3D:} total \textbf{train/test = 36{,}231/10{,}405}. ShapeNet contributes \textbf{35{,}708/10{,}261} CAD samples, and the semantically matched OmniObject3D subset contributes \textbf{523/144} real scans. Only OmniObject3D categories listed in Table~\ref{tab:omni2shape} are retained.

\item \textbf{ShapeNet + ShapeNet-C:} total \textbf{train/test = 35{,}708/20{,}522}. The training split contains \textbf{17{,}854} clean ShapeNet samples and \textbf{17{,}854} corrupted ShapeNet-C samples under a non-overlap object split. The corrupted branch is randomly sampled from the complete ShapeNet-C bank using only objects not selected for the clean branch. The test split contains the full clean ShapeNet test set \textbf{10{,}261} and a corresponding corrupted ShapeNet-C test subset of \textbf{10{,}261} samples.
\end{itemize}

\subsubsection{Semantic Mappings}
\label{app:mappings}

\noindent
Tables~\ref{tab:modelnet_mappings} and~\ref{tab:omni2shape} summarize the semantic bridges used to merge non-CAD datasets into the corresponding CAD taxonomies.

\begin{table*}[!t]
\centering
\small
\caption{Semantic alignments to ModelNet40 for ModelNet40 + ScanObjectNN and ModelNet40 + CO3D.}
\label{tab:modelnet_mappings}
\begin{minipage}{0.46\linewidth}
\centering
\textbf{(a) ScanObjectNN $\rightarrow$ ModelNet40}
\vspace{2mm}

\begin{tabular}{l|l}
\toprule
ScanObjectNN & ModelNet40 \\
\midrule
bed & bed \\
box & glass\_box \\
cabinet & wardrobe \\
chair & chair \\
desk & desk \\
display & monitor \\
door & door \\
shelf & bookshelf \\
sink & sink \\
sofa & sofa \\
table & table \\
toilet & toilet \\
\bottomrule
\end{tabular}
\end{minipage}
\hfill
\begin{minipage}{0.46\linewidth}
\centering
\textbf{(b) CO3D $\rightarrow$ ModelNet40}
\vspace{2mm}

\begin{tabular}{l|l}
\toprule
CO3D & ModelNet40 \\
\midrule
bench & bench \\
bottle & bottle \\
bowl & bowl \\
car & car \\
chair & chair \\
couch & sofa \\
cup & cup \\
keyboard & keyboard \\
laptop & laptop \\
plant & plant \\
toilet & toilet \\
plane & airplane \\
vase & vase \\
\bottomrule
\end{tabular}
\end{minipage}
\end{table*}

\begin{table*}[!t]
\centering
\small
\caption{Full OmniObject3D $\rightarrow$ ShapeNet55 name mapping used in ShapeNet + OmniObject3D.}
\label{tab:omni2shape}
\begin{tabular}{l|l||l|l}
\toprule
OmniObject3D & ShapeNet55 & OmniObject3D & ShapeNet55 \\
\midrule
bed & bed &
bottle & bottle \\
bowl & bowl &
cabinet & cabinet \\
chair & chair &
clock & clock \\
cup & mug &
guitar & guitar \\
hat & cap &
helmet & helmet \\
keyboard & keyboard &
knife & knife \\
laptop & laptop &
light & lamp \\
microwaveoven & microwave &
monitor & display \\
picnic\_basket & basket &
pillow & pillow \\
remote\_control & remote\_control &
skateboard & skateboard \\
sofa & sofa &
speaker & loudspeaker \\
table & table &
bus & bus \\
car & car &
motorcycle & motorcycle \\
plane & airplane &
train & train \\
\bottomrule
\end{tabular}
\end{table*}

\subsubsection{ShapeNet + ShapeNet-C Generation}
\label{app:shapenet_c_mix}

\paragraph{Training split.}
Let \(\mathcal{I}_{\mathrm{train}}\) be the ShapeNet training object index set.
We randomly partition \(\mathcal{I}_{\mathrm{train}}\)into two disjoint equal-size subsets,
\(\mathcal{I}_{\mathrm{clean}}\) and \(\mathcal{I}_{\mathrm{corr}}\), such that
\[
\mathcal{I}_{\mathrm{clean}} \cap \mathcal{I}_{\mathrm{corr}} = \emptyset,
\qquad
|\mathcal{I}_{\mathrm{clean}}| = |\mathcal{I}_{\mathrm{corr}}| = 17{,}854 .
\]
For indices in \(\mathcal{I}_{\mathrm{clean}}\), we keep the original clean ShapeNet point clouds.
For indices in \(\mathcal{I}_{\mathrm{corr}}\), we randomly sample corresponding ShapeNet-C point clouds of the same underlying objects from the complete ShapeNet-C bank.
When multiple corrupted candidates are available for an object, one candidate is randomly selected.
Therefore, each physical CAD object appears in exactly one training branch, either clean or corrupted, but never both.

\paragraph{Test split.}
For evaluation, we concatenate the full clean ShapeNet test split with a corresponding ShapeNet-C test subset.
Specifically, each of the \textbf{10{,}261} ShapeNet test objects appears once in the clean branch and once in the corrupted branch, giving \textbf{20{,}522} test samples in total.
When multiple corrupted candidates are available for a test object, one candidate is selected using the same fixed-seed selection protocol.
This design allows domain-wise evaluation between clean CAD point clouds and corrupted point clouds while preserving object-index correspondence.

\subsubsection{Data Availability and Licensing}
\label{app:license}

The source datasets used in this paper are available from their official providers, subject to their respective licenses or terms of use.
We do \textbf{not} redistribute any official dataset files, raw point clouds, reconstructed point clouds, meshes, or processed HDF5 shards derived from these datasets.
We will only provides the information needed to reproduce the benchmark construction after users obtain the official datasets themselves.
\textbf{Users are responsible for obtaining the source datasets from the official providers and complying with the corresponding licenses and terms of use.}
Thus, reproducing Domain3D-CIL requires following the mixed-benchmark protocol described above using properly obtained local copies of the source datasets.

\subsection{PTM-based CIL Experiment Details}
\label{app:std_and_normalized_gap}

In this subsection, we report the run-to-run standard deviations of the results in the main paper.
All entries are averaged over the same repeated runs used in Tab.~\ref{tab:ptm_results_full}, and are presented as mean \(\pm\) standard deviation.
For readability, we split the results by mixed benchmark:
ModelNet + ScanObjectNN in Tab.~\ref{tab:std_modelnet_scanobjectnn},
ModelNet + CO3D in Tab.~\ref{tab:std_modelnet_co3d},
ShapeNet + OmniObject3D in Tab.~\ref{tab:std_shapenet_omniobject3d},
and ShapeNet + ShapeNet-C in Tab.~\ref{tab:std_shapenet_shapenetc}.

We further clarify the cross-domain Last Accuracy gap used in the main paper.
For clarity and direct interpretability, the main paper reports the raw performance difference between the CAD domain and the corresponding real or corrupted domain:
\[
\mathrm{Gap}_{\mathrm{LA}}
=
\mathrm{LA}_{D_1} - \mathrm{LA}_{D_2},
\]
where \(\textcolor{cadcolor}{D_1}\) denotes the CAD domain and \(\textcolor{realcolor}{D_2}\) denotes the real or corrupted domain.
This raw gap directly reflects the observed cross-domain performance discrepancy after incremental learning.
As shown by the joint-training results~\ref{tab:ptm_results_full} in the main paper, the static gap between the two domains is relatively small compared with the gaps produced by most CIL methods, so the raw gap provides a simple and intuitive summary of the phenomenon.

For completeness, we also report a static-gap-normalized version of the LA Gap in this appendix.
Let \(\mathrm{LA}^{\mathrm{JT}}_{D_1}\) and \(\mathrm{LA}^{\mathrm{JT}}_{D_2}\) denote the joint-training Last Accuracy on the two domains of the same mixed benchmark.
We define
\(\mathrm{Gap}_{\mathrm{LA}} = \mathrm{LA}_{D_1} - \mathrm{LA}_{D_2}\) and
\(\mathrm{Gap}^{\mathrm{JT}}_{\mathrm{LA}} = \mathrm{LA}^{\mathrm{JT}}_{D_1} - \mathrm{LA}^{\mathrm{JT}}_{D_2}\).
The normalized LA Gap is then defined as the difference between the incremental-learning performance drops on the two domains:
\[
\mathrm{Gap}^{\dagger}_{\mathrm{LA}}
=
\left(\mathrm{LA}^{\mathrm{JT}}_{D_2} - \mathrm{LA}_{D_2}\right)
-
\left(\mathrm{LA}^{\mathrm{JT}}_{D_1} - \mathrm{LA}_{D_1}\right)
=
\mathrm{Gap}_{\mathrm{LA}} - \mathrm{Gap}^{\mathrm{JT}}_{\mathrm{LA}} .
\]
This quantity measures how much more the real or corrupted domain degrades than the CAD domain when moving from joint training to class-incremental learning.
A smaller \(\mathrm{Gap}^{\dagger}_{\mathrm{LA}}\) indicates that a method introduces less additional cross-domain imbalance beyond the joint-training reference.
Accordingly, the LA Gap columns in Tabs.~\ref{tab:std_modelnet_scanobjectnn}, \ref{tab:std_modelnet_co3d}, \ref{tab:std_shapenet_omniobject3d}, and~\ref{tab:std_shapenet_shapenetc} report \(\mathrm{Gap}^{\dagger}_{\mathrm{LA}}\), while the Cum. Acc. Gap columns retain the raw cumulative gap because joint training does not define an incremental cumulative-accuracy trajectory.

\begin{table*}[!t]
\centering
\caption{Mean and standard deviation on ModelNet + ScanObjectNN.}
\label{tab:std_modelnet_scanobjectnn}
\small
\resizebox{0.95\linewidth}{!}{
\begin{tabular}{l ccc ccc}
\toprule
\multirow{2}{*}{Method}
& \multicolumn{3}{c}{Last Acc.}
& \multicolumn{3}{c}{Cum. Acc.} \\
\cmidrule(lr){2-4} \cmidrule(lr){5-7}
& \textcolor{cadcolor}{$D_1$}$\!\uparrow$
& \textcolor{realcolor}{$D_2$}$\!\uparrow$
& \(\mathrm{Gap}^{\dagger}_{\mathrm{LA}}\!\downarrow\)
& \textcolor{cadcolor}{$D_1$}$\!\uparrow$
& \textcolor{realcolor}{$D_2$}$\!\uparrow$
& \(\mathrm{Gap}_{\mathrm{CA}}\!\downarrow\) \\
\midrule
Joint-Training
& \(\overline{93.68}\) & \(\overline{91.48}\) & \(\overline{0.00}\)
& - & - & - \\

L2P~\cite{wang2022learning}
& \(62.44_{\pm0.69}\) & \(52.49_{\pm1.06}\) & \(7.75_{\pm0.58}\)
& \(78.49_{\pm0.33}\) & \(72.54_{\pm0.82}\) & \(5.95_{\pm0.44}\) \\

DualPrompt~\cite{wang2022dualprompt}
& \(58.59_{\pm0.42}\) & \(45.92_{\pm1.24}\) & \(10.47_{\pm1.05}\)
& \(78.88_{\pm0.61}\) & \(74.65_{\pm0.70}\) & \(4.23_{\pm0.51}\) \\

CODA-Prompt~\cite{smith2023coda}
& \(61.14_{\pm0.83}\) & \(51.69_{\pm0.57}\) & \(7.25_{\pm0.91}\)
& \(77.51_{\pm0.29}\) & \(69.49_{\pm0.96}\) & \(8.02_{\pm1.18}\) \\

SLCA~\cite{zhang2023slca}
& \(85.05_{\pm0.44}\) & \(72.96_{\pm0.72}\) & \(9.89_{\pm0.53}\)
& \(92.13_{\pm0.18}\) & \(84.41_{\pm0.69}\) & \(7.72_{\pm0.84}\) \\

SimpleCIL~\cite{zhou2025revisiting}
& \(88.82_{\pm0.31}\) & \(72.37_{\pm1.10}\) & \(14.25_{\pm0.76}\)
& \(94.48_{\pm0.26}\) & \(80.29_{\pm0.73}\) & \(14.19_{\pm0.69}\) \\

APER~\cite{zhou2025revisiting}
& \(70.38_{\pm0.92}\) & \(35.79_{\pm1.38}\) & \(32.39_{\pm1.02}\)
& \(79.09_{\pm0.58}\) & \(53.19_{\pm1.21}\) & \(25.90_{\pm1.34}\) \\

EASE~\cite{zhou2024expandable}
& \(65.68_{\pm0.47}\) & \(52.88_{\pm0.91}\) & \(10.60_{\pm1.31}\)
& \(75.07_{\pm0.73}\) & \(70.65_{\pm0.44}\) & \(4.42_{\pm0.79}\) \\

RanPAC~\cite{mcdonnell2023ranpac}
& \(89.30_{\pm0.29}\) & \(80.52_{\pm0.67}\) & \(6.58_{\pm0.45}\)
& \(94.07_{\pm0.12}\) & \(88.70_{\pm0.51}\) & \(5.37_{\pm0.28}\) \\

MOS~\cite{sun2025mos}
& \(\underline{91.07_{\pm0.26}}\) & \(\underline{81.50_{\pm0.41}}\) & \(7.37_{\pm0.89}\)
& \(\underline{95.38_{\pm0.21}}\) & \(91.06_{\pm0.37}\) & \(4.32_{\pm0.53}\) \\

TUNA~\cite{wang2025integrating}
& \(88.82_{\pm0.18}\) & \(80.52_{\pm0.74}\) & \(\underline{6.10_{\pm0.32}}\)
& \(94.46_{\pm0.27}\) & \(\underline{91.69_{\pm0.46}}\) & \(\underline{2.77_{\pm0.44}}\) \\

\midrule
PolyMem (Ours)
& \(\textbf{92.50}_{\pm0.27}\) & \(\textbf{87.90}_{\pm0.55}\) & \(\textbf{2.40}_{\pm0.34}\)
& \(\textbf{95.90}_{\pm0.17}\) & \(\textbf{93.41}_{\pm0.42}\) & \(\textbf{2.48}_{\pm0.31}\) \\
\bottomrule
\end{tabular}
}
\end{table*}

\begin{table*}[!t]
\centering
\caption{Mean and standard deviation on ModelNet + CO3D.}
\label{tab:std_modelnet_co3d}
\small
\resizebox{0.95\linewidth}{!}{
\begin{tabular}{l ccc ccc}
\toprule
\multirow{2}{*}{Method}
& \multicolumn{3}{c}{Last Acc.}
& \multicolumn{3}{c}{Cum. Acc.} \\
\cmidrule(lr){2-4} \cmidrule(lr){5-7}
& \textcolor{cadcolor}{$D_1$}$\!\uparrow$
& \textcolor{realcolor}{$D_2$}$\!\uparrow$
& \(\mathrm{Gap}^{\dagger}_{\mathrm{LA}}\!\downarrow\)
& \textcolor{cadcolor}{$D_1$}$\!\uparrow$
& \textcolor{realcolor}{$D_2$}$\!\uparrow$
& \(\mathrm{Gap}_{\mathrm{CA}}\!\downarrow\) \\
\midrule
Joint-Training
& \(\overline{92.88}\) & \(\overline{89.21}\) & \(\overline{0.00}\)
& - & - & - \\

L2P~\cite{wang2022learning}
& \(52.03_{\pm1.18}\) & \(24.63_{\pm2.04}\) & \(23.73_{\pm1.47}\)
& \(68.21_{\pm0.77}\) & \(53.16_{\pm1.41}\) & \(15.04_{\pm1.09}\) \\

DualPrompt~\cite{wang2022dualprompt}
& \(52.07_{\pm0.96}\) & \(23.13_{\pm1.83}\) & \(25.27_{\pm2.22}\)
& \(72.07_{\pm0.49}\) & \(54.93_{\pm1.28}\) & \(17.14_{\pm1.53}\) \\

CODA-Prompt~\cite{smith2023coda}
& \(58.43_{\pm0.74}\) & \(36.51_{\pm1.62}\) & \(18.25_{\pm1.31}\)
& \(72.63_{\pm0.67}\) & \(55.20_{\pm1.05}\) & \(17.44_{\pm1.43}\) \\

SLCA~\cite{zhang2023slca}
& \(80.88_{\pm0.56}\) & \(58.41_{\pm1.14}\) & \(18.80_{\pm0.92}\)
& \(89.81_{\pm0.36}\) & \(77.23_{\pm0.93}\) & \(12.59_{\pm1.18}\) \\

SimpleCIL~\cite{zhou2025revisiting}
& \(83.14_{\pm0.47}\) & \(58.87_{\pm1.36}\) & \(20.60_{\pm1.04}\)
& \(90.93_{\pm0.41}\) & \(75.16_{\pm0.82}\) & \(15.77_{\pm1.23}\) \\

APER~\cite{zhou2025revisiting}
& \(55.31_{\pm1.09}\) & \(43.40_{\pm1.21}\) & \(8.24_{\pm0.86}\)
& \(72.35_{\pm0.70}\) & \(64.29_{\pm1.02}\) & \(8.07_{\pm0.74}\) \\

EASE~\cite{zhou2024expandable}
& \(49.39_{\pm0.84}\) & \(31.97_{\pm1.69}\) & \(13.75_{\pm1.84}\)
& \(65.64_{\pm0.91}\) & \(55.05_{\pm1.13}\) & \(10.59_{\pm0.96}\) \\

RanPAC~\cite{mcdonnell2023ranpac}
& \(\underline{90.34_{\pm0.35}}\) & \(\underline{81.32_{\pm0.79}}\) & \(\underline{5.35_{\pm0.48}}\)
& \(93.93_{\pm0.19}\) & \(85.37_{\pm0.64}\) & \(8.56_{\pm0.73}\) \\

MOS~\cite{sun2025mos}
& \(88.01_{\pm0.28}\) & \(76.51_{\pm0.66}\) & \(7.83_{\pm0.91}\)
& \(\textbf{94.38}_{\pm0.24}\) & \(\underline{87.00_{\pm0.49}}\) & \(\underline{7.38_{\pm0.44}}\) \\

TUNA~\cite{wang2025integrating}
& \(87.24_{\pm0.52}\) & \(66.10_{\pm1.07}\) & \(17.47_{\pm0.82}\)
& \(93.41_{\pm0.31}\) & \(79.74_{\pm0.88}\) & \(13.67_{\pm1.06}\) \\

\midrule
PolyMem (Ours)
& \(\textbf{90.67}_{\pm0.31}\) & \(\textbf{84.35}_{\pm0.68}\) & \(\textbf{2.65}_{\pm0.44}\)
& \(\underline{94.29_{\pm0.22}}\) & \(\textbf{88.56}_{\pm0.46}\) & \(\textbf{5.73}_{\pm0.39}\) \\
\bottomrule
\end{tabular}
}
\end{table*}

\begin{table*}[!t]
\centering
\caption{Mean and standard deviation on ShapeNet + OmniObject3D.}
\label{tab:std_shapenet_omniobject3d}
\small
\resizebox{0.95\linewidth}{!}{
\begin{tabular}{l ccc ccc}
\toprule
\multirow{2}{*}{Method}
& \multicolumn{3}{c}{Last Acc.}
& \multicolumn{3}{c}{Cum. Acc.} \\
\cmidrule(lr){2-4} \cmidrule(lr){5-7}
& \textcolor{cadcolor}{$D_1$}$\!\uparrow$
& \textcolor{realcolor}{$D_2$}$\!\uparrow$
& \(\mathrm{Gap}^{\dagger}_{\mathrm{LA}}\!\downarrow\)
& \textcolor{cadcolor}{$D_1$}$\!\uparrow$
& \textcolor{realcolor}{$D_2$}$\!\uparrow$
& \(\mathrm{Gap}_{\mathrm{CA}}\!\downarrow\) \\
\midrule
Joint-Training
& \(\overline{88.47}\) & \(\overline{86.11}\) & \(\overline{0.00}\)
& - & - & - \\

L2P~\cite{wang2022learning}
& \(57.08_{\pm1.31}\) & \(34.03_{\pm2.24}\) & \(20.69_{\pm1.73}\)
& \(75.58_{\pm0.86}\) & \(50.31_{\pm1.87}\) & \(25.27_{\pm2.10}\) \\

DualPrompt~\cite{wang2022dualprompt}
& \(59.61_{\pm1.05}\) & \(35.42_{\pm2.03}\) & \(21.83_{\pm2.37}\)
& \(79.54_{\pm0.79}\) & \(51.67_{\pm1.61}\) & \(27.87_{\pm1.92}\) \\

CODA-Prompt~\cite{smith2023coda}
& \(65.96_{\pm0.91}\) & \(44.44_{\pm1.74}\) & \(19.16_{\pm1.28}\)
& \(82.71_{\pm0.69}\) & \(61.87_{\pm1.30}\) & \(20.84_{\pm1.56}\) \\

SLCA~\cite{zhang2023slca}
& \(79.03_{\pm0.64}\) & \(59.03_{\pm1.47}\) & \(17.64_{\pm1.03}\)
& \(91.19_{\pm0.44}\) & \(71.64_{\pm1.12}\) & \(19.55_{\pm1.35}\) \\

SimpleCIL~\cite{zhou2025revisiting}
& \(77.52_{\pm0.78}\) & \(53.47_{\pm1.62}\) & \(21.69_{\pm1.84}\)
& \(88.98_{\pm0.52}\) & \(63.59_{\pm1.09}\) & \(25.39_{\pm1.47}\) \\

APER~\cite{zhou2025revisiting}
& \(68.52_{\pm1.26}\) & \(29.86_{\pm2.58}\) & \(36.30_{\pm2.06}\)
& \(80.28_{\pm0.94}\) & \(38.34_{\pm2.14}\) & \(41.94_{\pm1.73}\) \\

EASE~\cite{zhou2024expandable}
& \(58.74_{\pm0.87}\) & \(44.44_{\pm1.52}\) & \(11.94_{\pm1.96}\)
& \(75.29_{\pm0.74}\) & \(61.92_{\pm1.24}\) & \(13.37_{\pm0.91}\) \\

RanPAC~\cite{mcdonnell2023ranpac}
& \(83.26_{\pm0.48}\) & \(\underline{74.69_{\pm0.92}}\) & \(\underline{6.21_{\pm0.71}}\)
& \(89.98_{\pm0.33}\) & \(\underline{85.10_{\pm0.69}}\) & \(\underline{4.88_{\pm0.56}}\) \\

MOS~\cite{sun2025mos}
& \(81.95_{\pm0.37}\) & \(70.14_{\pm0.81}\) & \(9.45_{\pm1.02}\)
& \(92.29_{\pm0.29}\) & \(78.01_{\pm0.76}\) & \(14.28_{\pm0.67}\) \\

TUNA~\cite{wang2025integrating}
& \(\underline{84.48_{\pm0.53}}\) & \(65.28_{\pm1.11}\) & \(16.84_{\pm0.88}\)
& \(\textbf{93.20}_{\pm0.34}\) & \(76.53_{\pm0.91}\) & \(16.67_{\pm1.12}\) \\

\midrule
PolyMem (Ours)
& \(\textbf{87.15}_{\pm0.42}\) & \(\textbf{83.52}_{\pm0.86}\) & \(\textbf{1.27}_{\pm0.51}\)
& \(\underline{92.42_{\pm0.24}}\) & \(\textbf{89.63}_{\pm0.63}\) & \(\textbf{2.79}_{\pm0.47}\) \\
\bottomrule
\end{tabular}
}
\end{table*}

\begin{table*}[!t]
\centering
\caption{Mean and standard deviation on ShapeNet + ShapeNet-C.}
\label{tab:std_shapenet_shapenetc}
\small
\resizebox{0.95\linewidth}{!}{
\begin{tabular}{l ccc ccc}
\toprule
\multirow{2}{*}{Method}
& \multicolumn{3}{c}{Last Acc.}
& \multicolumn{3}{c}{Cum. Acc.} \\
\cmidrule(lr){2-4} \cmidrule(lr){5-7}
& \textcolor{cadcolor}{$D_1$}$\!\uparrow$
& \textcolor{realcolor}{$D_2$}$\!\uparrow$
& \(\mathrm{Gap}^{\dagger}_{\mathrm{LA}}\!\downarrow\)
& \textcolor{cadcolor}{$D_1$}$\!\uparrow$
& \textcolor{realcolor}{$D_2$}$\!\uparrow$
& \(\mathrm{Gap}_{\mathrm{CA}}\!\downarrow\) \\
\midrule
Joint-Training
& \(\overline{87.42}\) & \(\overline{85.01}\) & \(\overline{0.00}\)
& - & - & - \\

L2P~\cite{wang2022learning}
& \(61.96_{\pm0.88}\) & \(50.93_{\pm1.23}\) & \(8.62_{\pm0.74}\)
& \(80.22_{\pm0.61}\) & \(72.06_{\pm0.92}\) & \(8.16_{\pm1.05}\) \\

DualPrompt~\cite{wang2022dualprompt}
& \(62.69_{\pm0.73}\) & \(52.37_{\pm1.06}\) & \(7.91_{\pm1.18}\)
& \(77.15_{\pm0.66}\) & \(68.33_{\pm0.83}\) & \(8.82_{\pm0.69}\) \\

CODA-Prompt~\cite{smith2023coda}
& \(69.28_{\pm0.64}\) & \(56.98_{\pm0.94}\) & \(9.89_{\pm0.81}\)
& \(84.06_{\pm0.47}\) & \(74.72_{\pm0.71}\) & \(9.34_{\pm0.96}\) \\

SLCA~\cite{zhang2023slca}
& \(78.20_{\pm0.52}\) & \(63.35_{\pm0.81}\) & \(12.44_{\pm1.04}\)
& \(91.18_{\pm0.36}\) & \(82.00_{\pm0.58}\) & \(9.18_{\pm0.62}\) \\

SimpleCIL~\cite{zhou2025revisiting}
& \(76.60_{\pm0.39}\) & \(69.34_{\pm0.71}\) & \(4.85_{\pm0.43}\)
& \(88.50_{\pm0.31}\) & \(82.76_{\pm0.63}\) & \(5.74_{\pm0.55}\) \\

APER~\cite{zhou2025revisiting}
& \(66.75_{\pm0.81}\) & \(59.29_{\pm1.12}\) & \(5.05_{\pm0.77}\)
& \(79.11_{\pm0.58}\) & \(72.31_{\pm0.86}\) & \(6.80_{\pm0.91}\) \\

EASE~\cite{zhou2024expandable}
& \(68.82_{\pm0.79}\) & \(52.07_{\pm1.35}\) & \(14.34_{\pm1.11}\)
& \(84.79_{\pm0.69}\) & \(69.73_{\pm0.97}\) & \(15.06_{\pm1.26}\) \\

RanPAC~\cite{mcdonnell2023ranpac}
& \(76.33_{\pm0.34}\) & \(69.70_{\pm0.66}\) & \(\underline{4.21_{\pm0.37}}\)
& \(90.06_{\pm0.21}\) & \(84.34_{\pm0.52}\) & \(5.72_{\pm0.61}\) \\

MOS~\cite{sun2025mos}
& \(82.16_{\pm0.29}\) & \(72.29_{\pm0.62}\) & \(7.46_{\pm0.54}\)
& \(92.40_{\pm0.25}\) & \(86.53_{\pm0.46}\) & \(5.87_{\pm0.43}\) \\

TUNA~\cite{wang2025integrating}
& \(\underline{83.13_{\pm0.36}}\) & \(\underline{75.36_{\pm0.78}}\) & \(5.36_{\pm0.68}\)
& \(\underline{92.81_{\pm0.28}}\) & \(\underline{88.16_{\pm0.57}}\) & \(\underline{4.65_{\pm0.49}}\) \\

\midrule
PolyMem (Ours)
& \(\textbf{83.72}_{\pm0.24}\) & \(\textbf{77.85}_{\pm0.49}\) & \(\textbf{3.46}_{\pm0.41}\)
& \(\textbf{93.21}_{\pm0.19}\) & \(\textbf{89.45}_{\pm0.44}\) & \(\textbf{3.76}_{\pm0.36}\) \\
\bottomrule
\end{tabular}
}
\end{table*}

\begin{figure}[ht]
    \centering
    \begin{subfigure}[b]{0.49\linewidth}
        \centering
        \raisebox{0cm}{\includegraphics[width=\linewidth]{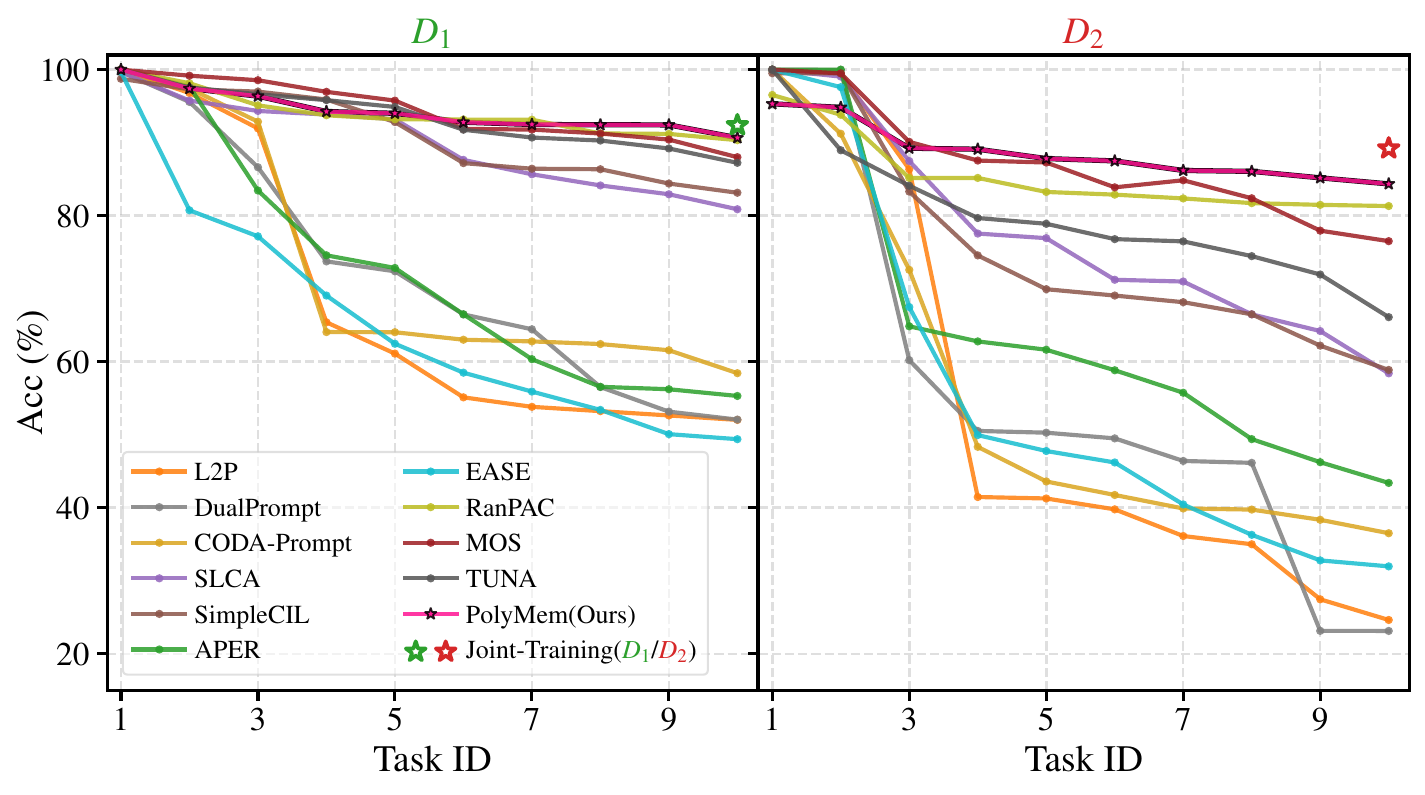}}
        \caption{ModelNet (\textcolor{cadcolor}{$D_1$}) + CO3D(\textcolor{realcolor}{$D_2$})}
        \label{fig:fig10a}
    \end{subfigure}%
    \hspace{0.0\linewidth}%
    \begin{subfigure}[b]{0.49\linewidth}
        \centering
        \raisebox{0cm}{\includegraphics[width=\linewidth]{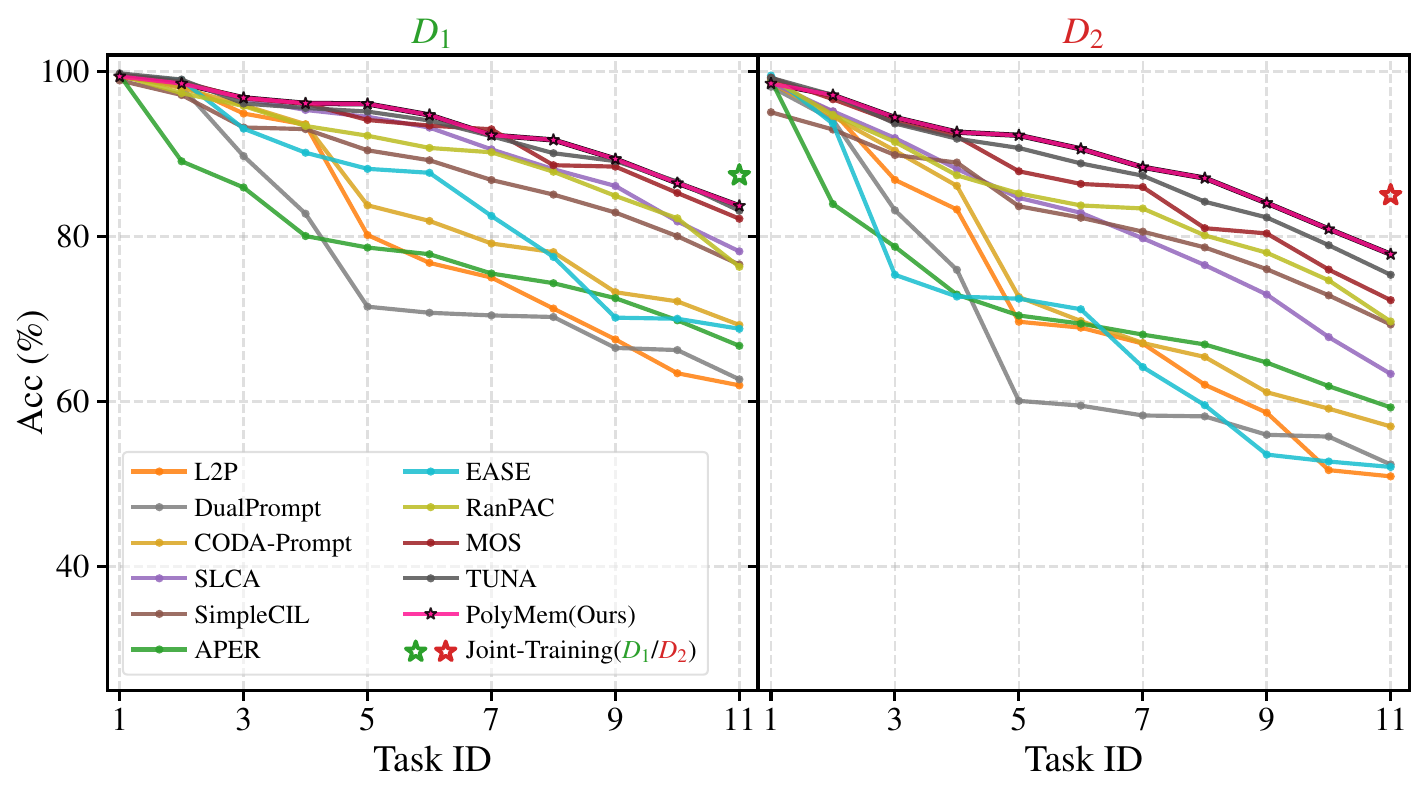}}
        \caption{ShapeNet (\textcolor{cadcolor}{$D_1$}) + ShapeNet-C (\textcolor{realcolor}{$D_2$})}
        \label{fig:fig10b}
    \end{subfigure}
    \caption{Task-wise forgetting curves showing performance discrepancy between $D_1$ and $D_2$.}
    \label{fig:fig10}
\end{figure}

The main paper reports task-wise forgetting curves for two representative mixed benchmarks, ModelNet + ScanObjectNN and ShapeNet + OmniObject3D in Fig.~\ref{fig:fig4_a} and Fig.~\ref{fig:fig4_b}.
Here, we provide the complementary curves for ModelNet + CO3D and ShapeNet + ShapeNet-C in Fig.~\ref{fig:fig10a} and Fig.~\ref{fig:fig10b}.

\begin{figure}[ht]
    \centering
    \begin{subfigure}[b]{0.49\linewidth}
        \centering
        \raisebox{0cm}{\includegraphics[width=\linewidth]{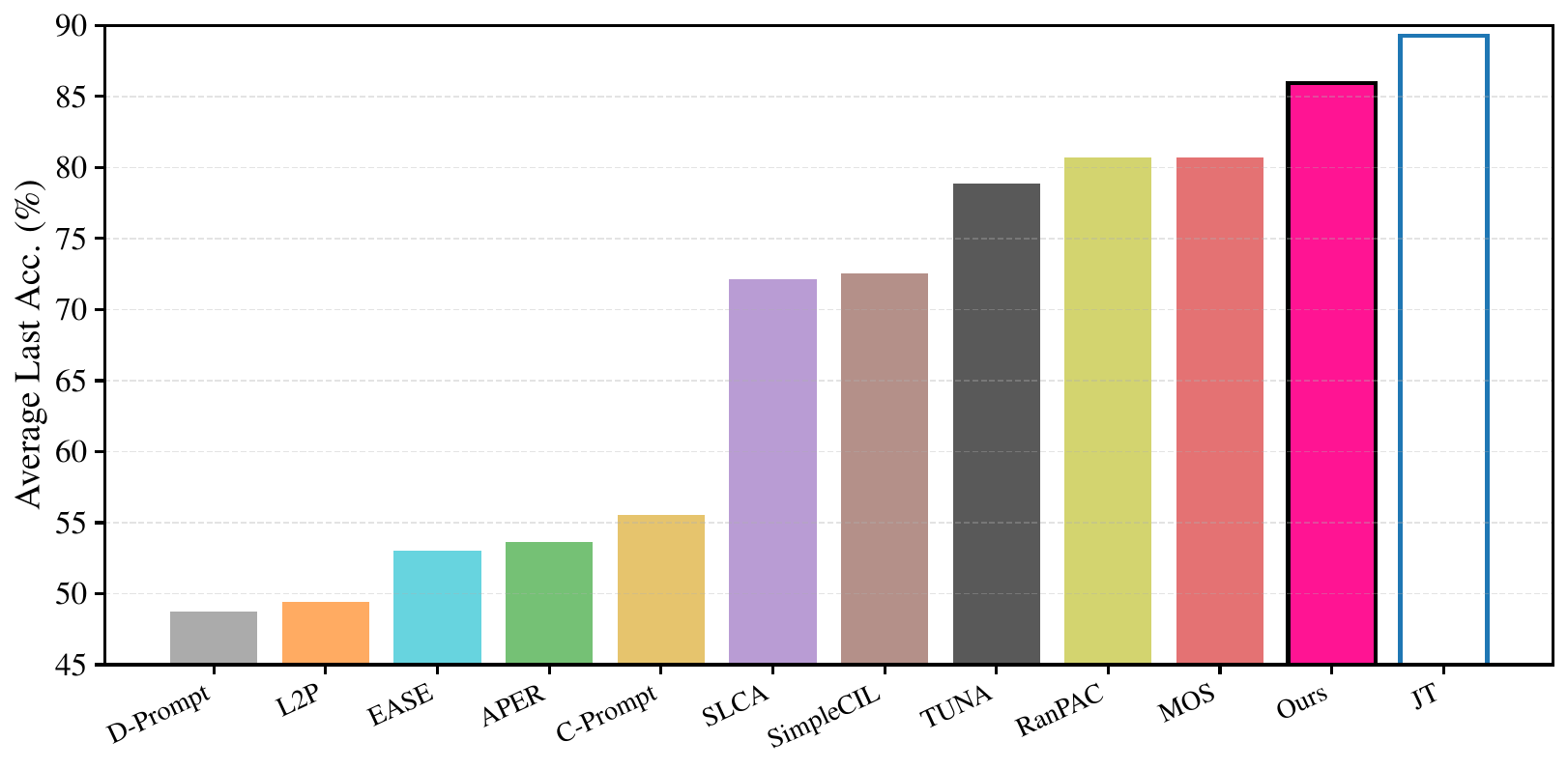}}
        \caption{}
        \label{fig:fig11a}
    \end{subfigure}%
    \hspace{0.0\linewidth}%
    \begin{subfigure}[b]{0.49\linewidth}
        \centering
        \raisebox{0cm}{\includegraphics[width=\linewidth]{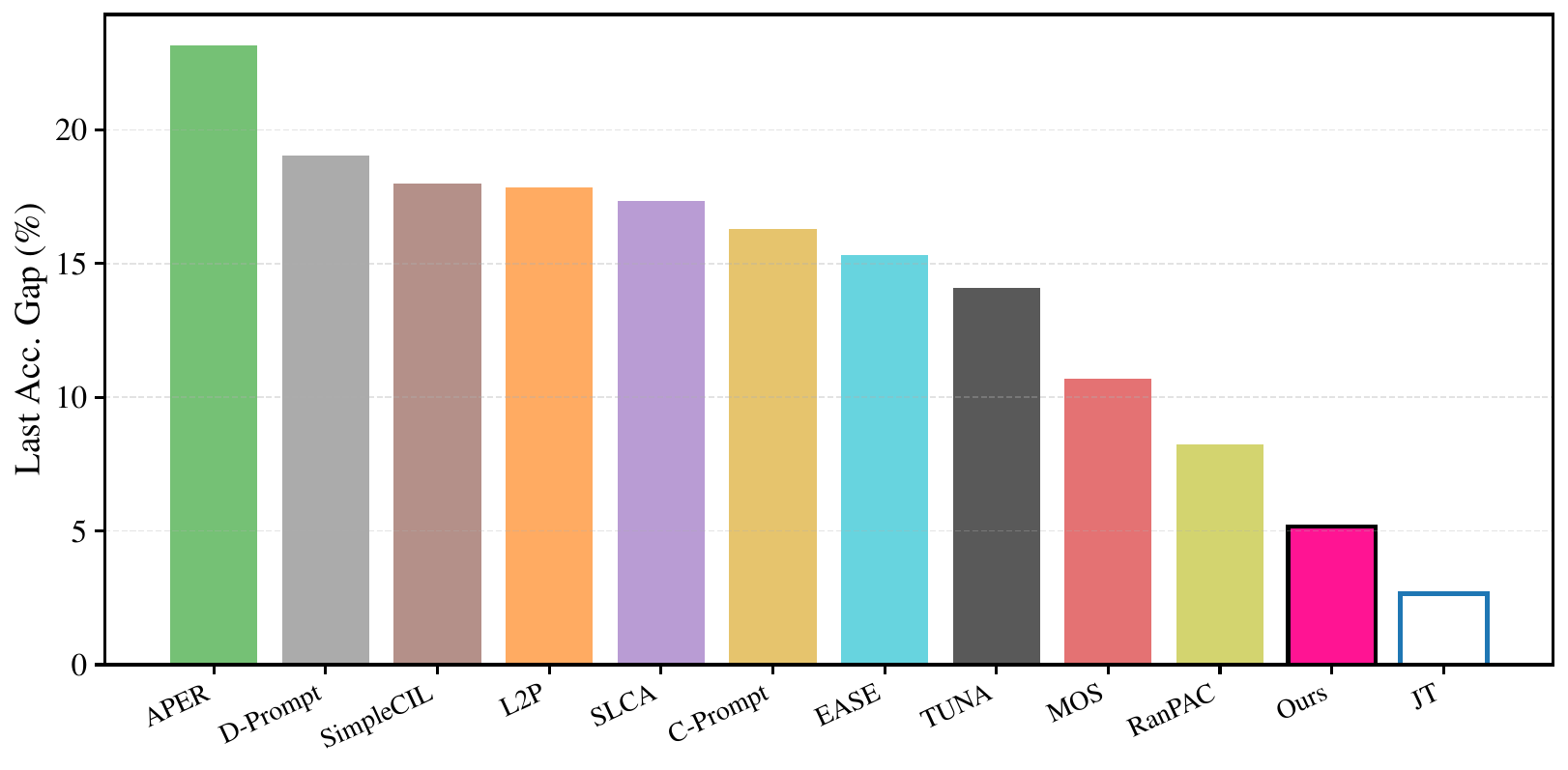}}
        \caption{}
        \label{fig:fig11b}
    \end{subfigure}
    \caption{
    \textbf{Overall comparison of PTM-based CIL methods on Domain3D-CIL.}
    Left: average Last Accuracy over both domains and all four mixed datasets.
    Right: average Last Accuracy Gap over the four mixed datasets.
    Higher average Last Accuracy and lower Last Accuracy Gap indicate better overall performance and stronger cross-domain balance.
    }
    \label{fig:fig11}
\end{figure}

Fig.~\ref{fig:fig11} summarizes the overall performance of different PTM-based CIL methods on the four mixed datasets of Domain3D-CIL.
For each method, we aggregate the final-task results over the two domains and over all mixed datasets, so as to jointly compare overall recognition performance and cross-domain balance.
Specifically, the left panel reports the cross-domain average Last Accuracy (LA), defined as
\[
\overline{\mathrm{LA}}
=
\frac{1}{|\mathcal{B}|}
\sum_{b \in \mathcal{B}}
\frac{
\mathrm{LA}^{b}_{D_1}
+
\mathrm{LA}^{b}_{D_2}
}{2},
\]
where \(\mathcal{B}\) denotes the set of four mixed datasets, and \(\textcolor{cadcolor}{D_1}\) and \(\textcolor{realcolor}{D_2}\) denote the CAD domain and the corresponding real or corrupted domain, respectively.
This metric measures the overall classification performance after all incremental tasks are completed.
The right panel reports the average LA Gap across the four mixed datasets:
\[
\overline{\mathrm{Gap}}_{\mathrm{LA}}
=
\frac{1}{|\mathcal{B}|}
\sum_{b \in \mathcal{B}}
\left(
\mathrm{LA}^{b}_{D_1}
-
\mathrm{LA}^{b}_{D_2}
\right),
\]
which measures the degree of cross-domain performance discrepancy.
An ideal method should achieve a high \(\overline{\mathrm{LA}}\) while maintaining a small \(\overline{\mathrm{Gap}}_{\mathrm{LA}}\).
As shown in Fig.~\ref{fig:fig11}, PolyMem achieves a favorable trade-off: it maintains strong overall performance while effectively reducing the cross-domain performance discrepancy.

\subsection{Additional Task-Split Experiments}
\label{app:additional_task_splits}

To further evaluate the robustness of PolyMem under different class-incremental task partitions, we conduct additional experiments with task splits that differ from the default settings used in the main paper.
Specifically, we consider two representative settings: ModelNet + CO3D with B20Inc4, where the base task contains 20 classes and each subsequent task introduces 4 new classes, resulting in 6 tasks in total; and ShapeNet + ShapeNet-C with B7Inc2, where the base task contains 7 classes and each subsequent task introduces 2 new classes, resulting in 25 tasks in total.
Compared with the default task splits, the former uses a larger base task and fewer incremental stages, while the latter uses a smaller increment size and many more incremental stages.
These two settings therefore provide complementary stress tests for examining whether the effectiveness of PolyMem depends on a particular task partition.

Tables~\ref{tab:add_modelnet_co3d_b20inc4} and~\ref{tab:add_shapenet_shapenetc_b7inc2} report the quantitative results under these two additional task splits.
The evaluation metrics follow the main experiments: we report Last Accuracy (LA), Cumulative Accuracy (CA), and the corresponding domain gap between the CAD domain \(\textcolor{cadcolor}{D_1}\) and the reconstructed or corrupted domain \(\textcolor{realcolor}{D_2}\).
Across both settings, the results are consistent with the observations in the main paper.
Existing PTM-based CIL baselines still exhibit clear performance discrepancy between the two domains, whereas PolyMem achieves strong performance on both domains and reduces the cross-domain gap.

\begin{table*}[!t]
\centering
\caption{Quantitative results on ModelNet + CO3D under the B20Inc4 task split.}
\label{tab:add_modelnet_co3d_b20inc4}
\small
\resizebox{0.95\linewidth}{!}{
\begin{tabular}{l ccc ccc}
\toprule
\multirow{2}{*}{Method}
& \multicolumn{3}{c}{Last Acc.}
& \multicolumn{3}{c}{Cum. Acc.} \\
\cmidrule(lr){2-4} \cmidrule(lr){5-7}
& \textcolor{cadcolor}{$D_1$}$\!\uparrow$
& \textcolor{realcolor}{$D_2$}$\!\uparrow$
& \(\mathrm{Gap}\!\downarrow\)
& \textcolor{cadcolor}{$D_1$}$\!\uparrow$
& \textcolor{realcolor}{$D_2$}$\!\uparrow$
& \(\mathrm{Gap}\!\downarrow\) \\
\midrule
Joint-Training
& \(\overline{92.88}\) & \(\overline{89.21}\) & \(\overline{3.67}\) & - & - & - \\

L2P~\cite{wang2022learning}
& \(64.29_{\pm0.72}\) & \(38.15_{\pm1.16}\) & \(26.14_{\pm0.93}\)
& \(74.44_{\pm0.48}\) & \(54.53_{\pm0.84}\) & \(19.91_{\pm0.71}\) \\

DualPrompt~\cite{wang2022dualprompt}
& \(70.06_{\pm0.61}\) & \(43.06_{\pm1.02}\) & \(27.00_{\pm1.28}\)
& \(79.49_{\pm0.44}\) & \(58.61_{\pm0.76}\) & \(20.88_{\pm0.65}\) \\

CODA-Prompt~\cite{smith2023coda}
& \(62.15_{\pm0.58}\) & \(54.82_{\pm0.95}\) & \(7.33_{\pm0.62}\)
& \(70.77_{\pm0.39}\) & \(65.23_{\pm0.68}\) & \(5.54_{\pm0.49}\) \\

SLCA~\cite{zhang2023slca}
& \(82.54_{\pm0.41}\) & \(69.89_{\pm0.86}\) & \(12.65_{\pm0.77}\)
& \(87.47_{\pm0.25}\) & \(74.25_{\pm0.58}\) & \(13.23_{\pm0.67}\) \\

SimpleCIL~\cite{zhou2025revisiting}
& \(82.98_{\pm0.33}\) & \(59.03_{\pm0.73}\) & \(23.95_{\pm0.96}\)
& \(86.24_{\pm0.29}\) & \(61.13_{\pm0.52}\) & \(25.11_{\pm0.74}\) \\

APER~\cite{zhou2025revisiting}
& \(75.08_{\pm0.63}\) & \(70.29_{\pm0.82}\) & \(\underline{4.79_{\pm0.55}}\)
& \(83.93_{\pm0.47}\) & \(83.80_{\pm0.69}\) & \(\textbf{0.14}_{\pm0.31}\) \\

EASE~\cite{zhou2024expandable}
& \(57.13_{\pm0.79}\) & \(28.98_{\pm1.08}\) & \(28.15_{\pm1.36}\)
& \(74.24_{\pm0.52}\) & \(59.88_{\pm0.74}\) & \(14.36_{\pm0.86}\) \\

RanPAC~\cite{mcdonnell2023ranpac}
& \(\underline{91.57_{\pm0.22}}\) & \(85.27_{\pm0.49}\) & \(6.30_{\pm0.42}\)
& \(\textbf{94.14}_{\pm0.16}\) & \(83.52_{\pm0.36}\) & \(10.61_{\pm0.31}\) \\

MOS~\cite{sun2025mos}
& \(91.30_{\pm0.28}\) & \(\underline{86.37_{\pm0.44}}\) & \(4.93_{\pm0.51}\)
& \(92.69_{\pm0.20}\) & \(\underline{88.08_{\pm0.31}}\) & \(4.61_{\pm0.37}\) \\

TUNA~\cite{wang2025integrating}
& \(89.02_{\pm0.37}\) & \(78.46_{\pm0.62}\) & \(10.56_{\pm0.58}\)
& \(92.36_{\pm0.24}\) & \(83.27_{\pm0.48}\) & \(9.09_{\pm0.45}\) \\

\midrule
PolyMem (Ours)
& \(\textbf{92.27}_{\pm0.21}\) & \(\textbf{88.28}_{\pm0.46}\) & \(\textbf{3.99}_{\pm0.33}\)
& \(\underline{93.69_{\pm0.15}}\) & \(\textbf{89.14}_{\pm0.38}\) & \(\underline{4.55_{\pm0.29}}\) \\
\bottomrule
\end{tabular}
}
\end{table*}

\begin{table*}[!t]
\centering
\caption{Quantitative results on ShapeNet + ShapeNet-C under the B7Inc2 task split.}
\label{tab:add_shapenet_shapenetc_b7inc2}
\small
\resizebox{0.95\linewidth}{!}{
\begin{tabular}{l ccc ccc}
\toprule
\multirow{2}{*}{Method}
& \multicolumn{3}{c}{Last Acc.}
& \multicolumn{3}{c}{Cum. Acc.} \\
\cmidrule(lr){2-4} \cmidrule(lr){5-7}
& \textcolor{cadcolor}{$D_1$}$\!\uparrow$
& \textcolor{realcolor}{$D_2$}$\!\uparrow$
& \(\mathrm{Gap}\!\downarrow\)
& \textcolor{cadcolor}{$D_1$}$\!\uparrow$
& \textcolor{realcolor}{$D_2$}$\!\uparrow$
& \(\mathrm{Gap}\!\downarrow\) \\
\midrule
Joint-Training
& \(\overline{87.42}\) & \(\overline{85.01}\) & \(\overline{2.41}\) & - & - & - \\

L2P~\cite{wang2022learning}
& \(42.35_{\pm1.42}\) & \(34.72_{\pm1.85}\) & \(7.63_{\pm1.22}\)
& \(62.33_{\pm0.94}\) & \(54.84_{\pm1.31}\) & \(7.48_{\pm0.88}\) \\

DualPrompt~\cite{wang2022dualprompt}
& \(45.33_{\pm1.26}\) & \(35.11_{\pm1.72}\) & \(10.22_{\pm1.85}\)
& \(65.83_{\pm0.82}\) & \(57.59_{\pm1.18}\) & \(8.24_{\pm1.27}\) \\

CODA-Prompt~\cite{smith2023coda}
& \(42.34_{\pm1.18}\) & \(33.56_{\pm1.58}\) & \(8.78_{\pm1.06}\)
& \(58.21_{\pm0.91}\) & \(49.03_{\pm1.09}\) & \(9.18_{\pm1.21}\) \\

SLCA~\cite{zhang2023slca}
& \(75.29_{\pm0.74}\) & \(57.00_{\pm1.28}\) & \(18.29_{\pm1.52}\)
& \(83.40_{\pm0.55}\) & \(70.18_{\pm0.96}\) & \(13.21_{\pm0.83}\) \\

SimpleCIL~\cite{zhou2025revisiting}
& \(76.27_{\pm0.62}\) & \(58.82_{\pm1.17}\) & \(17.45_{\pm1.34}\)
& \(83.91_{\pm0.46}\) & \(68.03_{\pm0.81}\) & \(15.88_{\pm1.05}\) \\

APER~\cite{zhou2025revisiting}
& \(64.24_{\pm1.07}\) & \(51.28_{\pm1.43}\) & \(12.96_{\pm1.16}\)
& \(75.26_{\pm0.71}\) & \(65.57_{\pm1.05}\) & \(9.69_{\pm0.79}\) \\

EASE~\cite{zhou2024expandable}
& \(28.71_{\pm1.58}\) & \(20.83_{\pm2.06}\) & \(7.88_{\pm1.48}\)
& \(48.20_{\pm1.12}\) & \(43.89_{\pm1.33}\) & \(\textbf{4.32}_{\pm0.94}\) \\

RanPAC~\cite{mcdonnell2023ranpac}
& \(81.78_{\pm0.53}\) & \(72.24_{\pm0.97}\) & \(9.54_{\pm0.74}\)
& \(88.63_{\pm0.34}\) & \(82.64_{\pm0.68}\) & \(5.98_{\pm0.59}\) \\

MOS~\cite{sun2025mos}
& \(\underline{84.99_{\pm0.49}}\) & \(\underline{75.90_{\pm0.88}}\) & \(9.09_{\pm0.96}\)
& \(\underline{90.41_{\pm0.31}}\) & \(\underline{83.79_{\pm0.59}}\) & \(6.62_{\pm0.46}\) \\

TUNA~\cite{wang2025integrating}
& \(81.76_{\pm0.66}\) & \(74.64_{\pm1.04}\) & \(\textbf{7.12}_{\pm0.81}\)
& \(87.72_{\pm0.42}\) & \(82.59_{\pm0.74}\) & \(5.13_{\pm0.52}\) \\

\midrule
PolyMem (Ours)
& \(\textbf{86.51}_{\pm0.38}\) & \(\textbf{78.85}_{\pm0.76}\) & \(\underline{7.66_{\pm0.62}}\)
& \(\textbf{91.40}_{\pm0.27}\) & \(\textbf{86.32}_{\pm0.58}\) & \(\underline{5.08_{\pm0.41}}\) \\
\bottomrule
\end{tabular}
}
\end{table*}

\begin{figure}[!t]
    \centering
    \begin{subfigure}[b]{0.326\linewidth}
        \centering
        \raisebox{0.01cm}{\includegraphics[width=\linewidth]{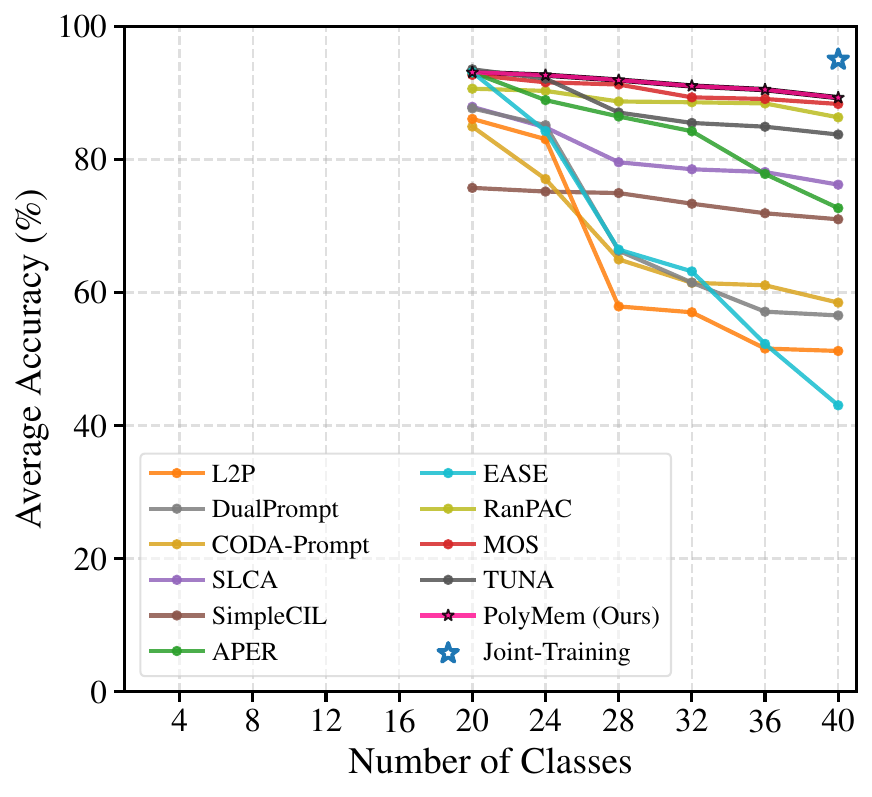}}
        \caption{ModelNet + CO3D, B20Inc4}
        \label{fig:fig12a}
    \end{subfigure}
    \hfill
    \begin{subfigure}[b]{0.665\linewidth}
        \centering
        \raisebox{0cm}{\includegraphics[width=\linewidth]{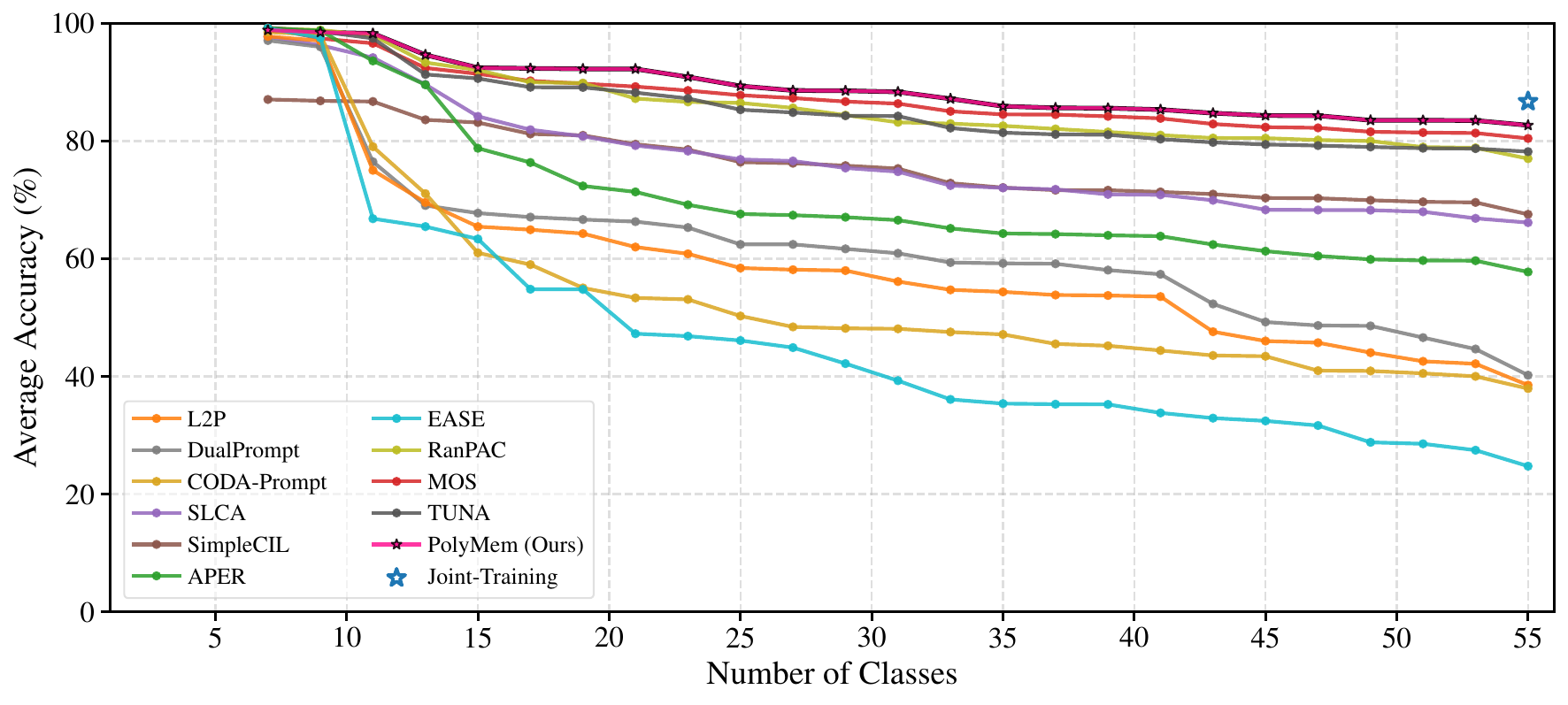}}
        \caption{ShapeNet + ShapeNet-C, B7Inc2}
        \label{fig:fig12b}
    \end{subfigure}
    \caption{
    Cross-domain average accuracy under additional task splits.
    Each curve reports the average accuracy over the CAD domain and the corresponding reconstructed or corrupted domain at each incremental task.
    Left: ModelNet + CO3D with B20Inc4 and 6 tasks.
    Right: ShapeNet + ShapeNet-C with B7Inc2 and 25 tasks.
    PolyMem maintains stronger average accuracy across the incremental process, confirming its robustness to different task-split designs.
    }
    \label{fig:fig12}
\end{figure}

Fig.~\ref{fig:fig12} further visualizes the task-wise cross-domain average accuracy under the two additional task splits.
To highlight the overall incremental learning performance, we report only one cross-domain curve for each method, obtained by averaging the two domain-specific accuracies at each incremental task.
After learning task \(t\), the model is evaluated on all classes observed so far.
We denote the corresponding accuracy on domain \(D_k\) as \(a^{t}_{D_k}\), where \(k\in\{1,2\}\).
The cross-domain average accuracy at task \(t\) is then defined as
\[
\mathrm{AvgAcc}_t
=
\frac{1}{2}
\left(
a^{t}_{D_1}
+
a^{t}_{D_2}
\right).
\]
This visualization focuses on the overall incremental learning performance under different task partitions, rather than separately showing the degradation on the two domains.

Fig.~\ref{fig:fig12a} shows ModelNet + CO3D under B20Inc4 with 6 tasks, while Fig.~\ref{fig:fig12b} shows ShapeNet + ShapeNet-C under B7Inc2 with 25 tasks.
Across both settings, PolyMem consistently maintains higher cross-domain average accuracy throughout the incremental process, especially in later tasks where catastrophic forgetting becomes more severe.
These results indicate that the advantage of PolyMem is not tied to the default task split used in the main paper, but remains stable under both fewer-task and many-task incremental settings.

\subsection{Parameter Analysis}
\label{app:param_storage_analysis}

We further analyze the number of learnable parameters and the auxiliary params cost of different PTM-based CIL methods.
Since all methods use the same frozen pre-trained backbone, we exclude the backbone parameters and report only the method-specific additional cost.
This analysis is conducted under the ShapeNet + ShapeNet-C B7Inc2 setting corresponding to Fig.~\ref{fig:fig12b}.

We adopt a two-column accounting protocol.
The first column, \textbf{Learnable Params}, counts the accumulated newly introduced learnable parameters, such as prompts, adapters, or classifier-related learnable parameters.
The second column, \textbf{Auxiliary Params}, counts non-backbone auxiliary params; these entries are not trainable parameters, but are explicitly stored and involved in computation, such as feature statistics or covariance-like matrices.
For methods that keep a complete copy or adapter-side reference of the Uni3D-Base feature extractor, the corresponding auxiliary params is counted as \(88.961M\), which is the parameter count of Uni3D-Base under our implementation.
All values are reported in \(M\), where \(1M = 10^6\) scalar entries.

For statistical-memory methods, the main auxiliary params comes from the accumulated matrices used for closed-form classification.
Similar to RanPAC, PolyMem stores
\[
G \in \mathbb{R}^{M \times M},
\qquad
Q \in \mathbb{R}^{M \times C_{\mathrm{tot}}},
\]
where \(M=10000\) is the lifted feature dimension and \(C_{\mathrm{tot}}=55\) is the total number of classes in the ShapeNet-based setting.
Therefore, \(G+Q\) contains
\[
M^2 + M C_{\mathrm{tot}} = 100.550M
\]
scalar entries.

In addition, PolyMem uses fixed TensorSketch descriptors to construct polynomial lifted features.
With Taylor order \(L=4\) and backbone feature dimension \(D=1024\), the hash indices and random signs occupy
\[
2 \times \frac{L(L+1)}{2}D
=
2 \times 10 \times 1024
=
20480
\approx 0.021M
\]
scalar entries in total.
Thus, the total auxiliary params of PolyMem is
\[
100.550M + 0.021M = 100.571M.
\]

\begin{table}[!t]
\centering
\caption{Learnable parameters and auxiliary params of PTM-based CIL methods. The frozen pre-trained backbone is excluded. All values are reported in \(M\), where \(1M=10^6\) scalar entries.}
\label{tab:param_storage_analysis}
\resizebox{0.6\linewidth}{!}{
\begin{tabular}{lcc}
\toprule
Method
& Learnable Params (M)
& Auxiliary Params (M) \\
\midrule
L2P~\cite{wang2022learning}
& 1.88 & 88.96 \\
DualPrompt~\cite{wang2022dualprompt}
& 0.97 & 88.96 \\
CODA-Prompt~\cite{smith2023coda}
& 3.10 & -- \\
SLCA~\cite{zhang2023slca}
& 0.05 & 32.44 \\
SimpleCIL~\cite{zhou2025revisiting}
& 0.05 & -- \\
MOS~\cite{sun2025mos}
& 7.34 & 32.44 \\
TUNA~\cite{wang2025integrating}
& 7.65 & 32.44 \\
RanPAC~\cite{mcdonnell2023ranpac}
& 0.53 & 111.04 \\
APER~\cite{zhou2025revisiting}
& 89.07 & 88.96 \\
EASE~\cite{zhou2024expandable}
& 0.98 & -- \\
PolyMem (Ours)
& 0.53 & 100.57 \\
\bottomrule
\end{tabular}
}
\end{table}

As shown in Tab.~\ref{tab:param_storage_analysis}, PolyMem introduces the same number of additional learnable parameters as RanPAC under this accounting protocol, since both methods rely on a frozen backbone and a closed-form classifier in the lifted feature space.
The main auxiliary params cost of PolyMem comes from the accumulated statistical memory \(G+Q\), while the TensorSketch descriptors add only \(0.021M\) scalar entries, which is negligible compared with the \(M^2\)-scale matrix \(G\).
Therefore, PolyMem improves the expressiveness of statistical memory through structured polynomial lifting without introducing substantial additional parameter or storage overhead.

\subsection{Experiment with Different Backbones}
\label{app:backbone_generalization}

To further verify that the effectiveness of PolyMem is not specific to the Uni3D-Base backbone used in the main experiments, we conduct additional experiments with two publicly available 3D pre-trained backbones: Point-BERT~\cite{yu2022point} and Point-MAE~\cite{pang2023masked}.
These two backbones follow different 3D pre-training paradigms from Uni3D, and therefore provide complementary evidence for evaluating the backbone generality of PolyMem.

We evaluate the two backbones on two object-level Domain3D-CIL mixed benchmarks: ModelNet + ScanObjectNN and ModelNet + CO3D.
For each backbone and each mixed benchmark, we report four representative settings: Joint-Training, SimpleCIL, RanPAC, and PolyMem.
Joint-Training serves as the non-incremental reference, while SimpleCIL and RanPAC are selected as representative statistical-memory baselines.
SimpleCIL stores first-order class prototypes, RanPAC performs random lifted-space ridge regression, and PolyMem replaces the random lifting with structured polynomial feature interactions while following the same replay-free PTM-based CIL protocol.

Tab.~\ref{tab:backbone_generalization} reports the results.
Although the absolute performance varies across different pre-trained backbones, the performance discrepancy phenomenon consistently appears: the accuracy on \(\textcolor{realcolor}{D_2}\) is generally lower than that on \(\textcolor{cadcolor}{D_1}\), especially after class-incremental learning.
More importantly, PolyMem consistently improves over RanPAC under both Point-BERT and Point-MAE, showing higher accuracy on the non-CAD domain and a smaller cross-domain gap.
These results indicate that PolyMem is not tied to a particular backbone architecture or pre-training recipe, but provides a general feature-space statistical modeling strategy for mitigating performance discrepancy in Domain3D-CIL.

\begin{table*}[!t]
\centering
\caption{Backbone generalization analysis on Domain3D-CIL.
We evaluate Point-BERT and Point-MAE on ModelNet + ScanObjectNN and ModelNet + CO3D.
\(\textcolor{cadcolor}{D_1}\) denotes the CAD domain and \(\textcolor{realcolor}{D_2}\) denotes the real-scanned or reconstructed domain.}
\label{tab:backbone_generalization}

\resizebox{1\linewidth}{!}{
\begin{tabular}{l ccc ccc ccc ccc}
\toprule
\multirow{3}{*}{Method}
  & \multicolumn{6}{c}{ModelNet (\textcolor{cadcolor}{$D_1$}) + ScanObjectNN (\textcolor{realcolor}{$D_2$})}
  & \multicolumn{6}{c}{ModelNet (\textcolor{cadcolor}{$D_1$}) + CO3D (\textcolor{realcolor}{$D_2$})} \\
\cmidrule(lr){2-7} \cmidrule(lr){8-13}
  & \multicolumn{3}{c}{Last Acc.}
  & \multicolumn{3}{c}{Cum. Acc.}
  & \multicolumn{3}{c}{Last Acc.}
  & \multicolumn{3}{c}{Cum. Acc.} \\
\cmidrule(lr){2-4} \cmidrule(lr){5-7} \cmidrule(lr){8-10} \cmidrule(lr){11-13}
  & \textcolor{cadcolor}{$D_1$}$\!\uparrow$
  & \textcolor{realcolor}{$D_2$}$\!\uparrow$
  & $\mathrm{Gap}\!\downarrow$
  & \textcolor{cadcolor}{$D_1$}$\!\uparrow$
  & \textcolor{realcolor}{$D_2$}$\!\uparrow$
  & $\mathrm{Gap}\!\downarrow$
  & \textcolor{cadcolor}{$D_1$}$\!\uparrow$
  & \textcolor{realcolor}{$D_2$}$\!\uparrow$
  & $\mathrm{Gap}\!\downarrow$
  & \textcolor{cadcolor}{$D_1$}$\!\uparrow$
  & \textcolor{realcolor}{$D_2$}$\!\uparrow$
  & $\mathrm{Gap}\!\downarrow$ \\
\midrule
\multicolumn{13}{c}{\textbf{Point-BERT Backbone}} \\
\midrule
Joint-Training
            & $\overline{91.45}$ & $\overline{82.52}$ & $\overline{8.93}$
            & - & - & -
            & $\overline{91.53}$ & $\overline{84.76}$ & $\overline{6.77}$
            & - & - & - \\

SimpleCIL~\cite{zhou2025revisiting}
            & 90.64 & 37.97 & 52.67
            & 94.82 & 44.33 & 50.49
            & 89.26 & 25.94 & 63.32
            & 93.88 & 33.31 & 60.57 \\

RanPAC~\cite{mcdonnell2023ranpac}
            & 90.93 & 66.38 & 24.55
            & 94.79 & 79.14 & 15.65
            & 90.24 & 65.07 & 25.17
            & 93.04 & 71.26 & 21.78 \\

PolyMem (Ours)
            & \textbf{91.00} & \textbf{73.96} & \textbf{17.04}
            & \textbf{95.02} & \textbf{83.04} & \textbf{11.98}
            & \textbf{91.53} & \textbf{70.39} & \textbf{21.14}
            & \textbf{94.11} & \textbf{75.06} & \textbf{19.05} \\

\specialrule{0.12em}{0.20em}{0.10em}
\multicolumn{13}{c}{\textbf{Point-MAE Backbone}} \\
\midrule
Joint-Training
            & $\overline{90.84}$ & $\overline{86.72}$ & $\overline{4.12}$
            & - & - & -
            & $\overline{90.73}$ & $\overline{86.39}$ & $\overline{4.34}$
            & - & - & - \\

SimpleCIL~\cite{zhou2025revisiting}
            & 77.48 & 65.23 & 12.25
            & 87.91 & 77.91 & 10.00
            & 62.44 & 39.23 & 23.21
            & 71.62 & 48.95 & 22.67 \\

RanPAC~\cite{mcdonnell2023ranpac}
            & 88.19 & 77.93 & 10.26
            & 92.66 & 82.29 & 10.37
            & 88.82 & 73.67 & 15.15
            & 92.71 & 78.35 & 14.36 \\

PolyMem (Ours)
            & \textbf{90.60} & \textbf{83.30} & \textbf{7.30}
            & \textbf{94.54} & \textbf{87.72} & \textbf{6.82}
            & \textbf{89.06} & \textbf{76.87} & \textbf{12.19}
            & \textbf{93.83} & \textbf{81.07} & \textbf{12.76} \\

\bottomrule
\end{tabular}
}
\end{table*}

\subsection{Compute Resources}
\label{app:compute_resources}

The running-time comparison in Fig.~\ref{fig:fig9} and the parameter analysis in Tab.~\ref{tab:param_storage_analysis} are measured under the same computational environment described below.
All experiments were conducted on compute nodes equipped with \(8\times\) NVIDIA A40 GPUs.
Each GPU provides 46,068 MiB of memory, i.e., about 48 GB.
The NVIDIA driver version is 570.195.03, and the CUDA compatibility version reported by the driver is 12.8.
This reported CUDA version indicates the highest CUDA toolkit level supported by the driver, and does not necessarily correspond to the CUDA version used to compile a specific deep learning framework.

For each random seed, a single training run uses only one GPU.
We do not use multi-GPU data parallelism within a single run.
Multiple GPUs on the same node are used only to execute different experiments in parallel, such as different methods, random seeds, or hyperparameter configurations.

\subsection{Classic CIL Paradigm}
\label{app:classic_cil_3d}

In addition to the PTM-based CIL experiments in Tab.~\ref{tab:ptm_results_full}, as shown in Fig.~\ref{fig:fig1_b}, we also evaluate a set of classic CIL baselines on Domain3D-CIL.
These methods are adapted to the 3D setting by replacing the 2D backbone with PointNet++ and training the model from scratch on point clouds.
Following the common replay-based protocol for classic CIL methods, we use an exemplar memory budget of 10 samples per class.
Since these methods rely on from-scratch training and exemplar replay, while the main experiments follow the PTM-based replay-free paradigm with a frozen Uni3D backbone, the two groups of methods are not directly comparable under a strictly fair protocol.
Therefore, we report the classic CIL results in the appendix rather than including them in the main experimental comparison.

Tab.~\ref{tab:classic_results_full} reports the results on the object-level Domain3D-CIL mixed benchmarks.
We use the same notation as in the main experiments: \(\textcolor{cadcolor}{D_1}\) denotes the CAD domain, \(\textcolor{realcolor}{D_2}\) denotes the corresponding real-scanned or reconstructed domain, and \(\mathrm{Gap}=D_1-D_2\) measures the cross-domain performance discrepancy.
Compared with the results in Tab.~\ref{tab:ptm_results_full}, the joint-training performance of PointNet++ without pre-training is slightly lower than that of pre-trained Uni3D on both domains, especially on \(\textcolor{realcolor}{D_2}\).
The static performance gap under joint training is also slightly larger, around \(4\%\)--\(5\%\).
Moreover, the performance discrepancy induced by CIL becomes more pronounced.
Unlike the PTM-based setting, where strong baselines such as RanPAC, MOS, and TUNA can partially alleviate the gap, most classic CIL baselines exhibit Last Acc. and Cum. Acc. gaps above \(20\%\).

Although classic CIL methods follow a different training paradigm, the results still show a clear and consistent performance discrepancy: the accuracy on \(\textcolor{realcolor}{D_2}\) is usually much lower than that on \(\textcolor{cadcolor}{D_1}\), and compared with joint training, the accuracy drop on \(\textcolor{realcolor}{D_2}\) is substantially larger than that on \(\textcolor{cadcolor}{D_1}\).
This indicates that CIL baselines without pre-training suffer from more severe catastrophic forgetting and performance discrepancy.

These results suggest that the performance discrepancy phenomenon is not specific to PTM-based CIL methods.
When heterogeneous 3D domains are mixed within shared semantic classes, the same issue also appears in classic CIL.
Future work may further investigate this more challenging setting.

\begin{table*}[!ht]
\centering
\caption{Quantitative evaluation of classic CIL baselines on Domain3D-CIL.
All methods use PointNet++ trained from scratch with a replay budget of 10 exemplars per class.}
\label{tab:classic_results_full}

\resizebox{1\linewidth}{!}{
\begin{tabular}{l ccc ccc ccc ccc}
\toprule
\multirow{3}{*}{Method}
  & \multicolumn{6}{c}{ModelNet (\textcolor{cadcolor}{$D_1$}) + ScanObjectNN (\textcolor{realcolor}{$D_2$})}
  & \multicolumn{6}{c}{ModelNet (\textcolor{cadcolor}{$D_1$}) + CO3D (\textcolor{realcolor}{$D_2$})} \\
\cmidrule(lr){2-7} \cmidrule(lr){8-13}
  & \multicolumn{3}{c}{Last Acc.}
  & \multicolumn{3}{c}{Cum. Acc.}
  & \multicolumn{3}{c}{Last Acc.}
  & \multicolumn{3}{c}{Cum. Acc.} \\
\cmidrule(lr){2-4} \cmidrule(lr){5-7} \cmidrule(lr){8-10} \cmidrule(lr){11-13}
  & \textcolor{cadcolor}{$D_1$}$\!\uparrow$
  & \textcolor{realcolor}{$D_2$}$\!\uparrow$
  & $\mathrm{Gap}\!\downarrow$
  & \textcolor{cadcolor}{$D_1$}$\!\uparrow$
  & \textcolor{realcolor}{$D_2$}$\!\uparrow$
  & $\mathrm{Gap}\!\downarrow$
  & \textcolor{cadcolor}{$D_1$}$\!\uparrow$
  & \textcolor{realcolor}{$D_2$}$\!\uparrow$
  & $\mathrm{Gap}\!\downarrow$
  & \textcolor{cadcolor}{$D_1$}$\!\uparrow$
  & \textcolor{realcolor}{$D_2$}$\!\uparrow$
  & $\mathrm{Gap}\!\downarrow$ \\
\midrule
Joint-Training
& \(\overline{90.72}\) & \(\overline{86.51}\) & \(\overline{4.21}\) & - & - & -
& \(\overline{91.21}\) & \(\overline{86.62}\) & \(\overline{4.59}\) & - & - & - \\

ER
& 69.00 & 40.95 & 28.05 & 74.17 & 46.38 & 27.79
& 66.17 & 35.28 & 30.89 & 67.53 & 38.53 & 29.00 \\

iCaRL~\cite{rebuffi2017icarl}
& 72.20 & 39.88 & 32.32 & 79.29 & 53.63 & 25.66
& 65.15 & 39.41 & 25.74 & 72.66 & 48.99 & 23.67 \\

BiC~\cite{wu2019large}
& 60.94 & 38.57 & 22.37 & 63.60 & 40.67 & 22.93
& 61.39 & 38.46 & 22.93 & 61.39 & 42.92 & 18.47 \\

WA~\cite{zhao2020maintaining}
& 71.03 & 38.67 & 32.36 & 79.18 & 51.73 & 27.45
& 68.68 & 46.17 & 22.51 & 75.77 & 52.87 & 22.90 \\

PODNet~\cite{douillard2020podnet}
& 68.68 & 38.17 & 30.51 & 80.13 & 52.04 & 28.09
& 62.93 & 36.33 & 26.60 & 79.10 & 51.08 & 28.02 \\

DER~\cite{yan2021dynamically}
& 74.59 & 44.33 & 30.26 & 83.99 & 56.60 & 27.39
& 67.10 & 55.74 & 11.36 & 78.41 & 61.81 & 16.60 \\

FOSTER~\cite{wang2022foster}
& 65.19 & 43.94 & 21.25 & 72.18 & 50.93 & 21.25
& 64.14 & 44.08 & 20.06 & 75.37 & 50.23 & 25.14 \\

iNeMo~\cite{fischer2024inemo}
& 69.98 & 35.39 & 34.59 & 76.88 & 46.08 & 30.80
& 57.74 & 42.49 & 15.25 & 63.82 & 46.30 & 17.52 \\
\bottomrule
\end{tabular}
}
\end{table*}

\subsection{Baseline Implementation Details}
\label{app:baseline_implementation}

The PTM-based CIL baselines in the main paper are implemented by adapting the official LAMDA-PILOT codebase~\cite{zhou2024continual} to the 3D setting. Since Uni3D~\cite{zhouuni3d} follows the same architecture with ViT , its interface is largely compatible with the vision-transformer backbones used in LAMDA-PILOT. Therefore, most baselines require only minimal modifications, mainly replacing the original 2D image encoder with the frozen Uni3D point-cloud encoder and adapting the data loader, feature extraction, and evaluation pipeline to point-cloud inputs. We keep the core algorithmic components of each baseline unchanged whenever possible, including prompt selection, prototype/statistical memory construction, classifier update rules, and evaluation metrics. The \textbf{incomplete} implementation of the PTM-based CIL baselines used for review is included in the \textbf{supplementary materials}. The complete and cleaned codebase including the proposed method \textbf{PolyMem} will be publicly released upon acceptance.

\end{document}